\documentclass[lettersize,journal]{IEEEtran}
\usepackage{amsmath,amsfonts}
\usepackage{algorithmic}
\usepackage{algorithm}
\usepackage{array}
\usepackage[caption=false,font=normalsize,labelfont=sf,textfont=sf]{subfig}
\usepackage{textcomp}
\usepackage{stfloats}
\usepackage{url}
\usepackage{verbatim}
\usepackage{graphicx}
\usepackage{cite}
\usepackage{tcolorbox}
\usepackage{array}
\usepackage{multirow}
\usepackage{graphicx}
\usepackage{balance}
\usepackage{booktabs}
\definecolor{mblue}{RGB}{0, 77, 128}
\usepackage{xcolor}    
\usepackage{colortbl}
\usepackage{arydshln}
\usepackage{multirow}
\usepackage{pifont}
\begin{document}

\title{Quality Assessment for AI Generated Images with Instruction Tuning}

\author{
Jiarui Wang$^{1}$, \qquad 
Huiyu Duan$^{1,2}$, \qquad 
Guangtao Zhai$^{1,2}$, \qquad
Xiongkuo Min$^{1*}$,\qquad \thanks{$^{*}$Corresponding Author.}
\\
$^{1}$Institute of Image Communication and Network Engineering, \\
$^{2}$ MoE Key Lab of Artificial Intelligence, AI Institute,\\ Shanghai Jiao Tong University, Shanghai, China\\}

% \institute{Institute of Image Communication and Network Engineering, \\
% \and MoE Key Lab of Artificial Intelligence, AI Institute,\\ Shanghai Jiao Tong University, Shanghai, China\\
% \and Tianjin University, Tianjin, China\\
% \and Shanghai Polytechnic University, Shanghai, China
% \email{\{wangjiarui,huiyuduan,minxiongkuo,zhaiguangtao\}@sjtu.edu.cn, jliu\_tju@tju.edu.cn, chenshi@sspu.edu.cn }}

% \IEEEpubid{0000--0000/00\$00.00~\copyright~2021 IEEE}
% Remember, if you use this you must call \IEEEpubidadjcol in the second
% column for its text to clear the IEEEpubid mark.

\maketitle
\begin{abstract}
Artificial Intelligence Generated Content (AIGC) has grown rapidly in recent years, among which AI-based image generation has gained widespread attention due to its efficient and imaginative image creation ability.
 However, AI-generated
Images (AIGIs) may not satisfy human preferences due to their unique distortions, which highlights the necessity to understand and evaluate human preferences for AIGIs.
To this end, in this paper, we first establish a novel Image Quality Assessment (IQA) database for AIGIs, termed AIGCIQA2023+, which provides human visual preference scores and detailed preference explanations from three perspectives including quality, authenticity, and correspondence.
Then, based on the constructed AIGCIQA2023+ database, this paper presents a \textbf{MINT-IQA} model to evaluate and explain human preferences for AIGIs from \underline{M}ulti-perspectives with \underline{IN}struction \underline{T}uning. 
Specifically, the MINT-IQA model first learn and evaluate human preferences for AI-generated Images from multi-perspectives, then via the vision-language instruction tuning strategy, MINT-IQA attains powerful understanding and explanation ability for human visual preference on AIGIs, which can be used for feedback to further improve the assessment capabilities.
Extensive experimental results demonstrate that the proposed MINT-IQA model achieves state-of-the-art performance in understanding and evaluating human visual preferences for AIGIs, and the proposed model also achieves competing results on traditional IQA tasks compared with state-of-the-art IQA models.  
The AIGCIQA2023+ database and MINT-IQA model are available at: https://github.com/IntMeGroup/MINT-IQA
\end{abstract}


\section{Introduction}

\label{sec:intro}
Artificial Intelligence Generated Content (AIGC) refers to the content generated with the assistance of AI, including texts, images, audios, \textit{etc.}
As an important part of AIGC, AI Generated Images (AIGIs) have gained significant attention due to their broad application prospects, thus, many AI-based image generation methods have been proposed in recent years, such as DALLE \cite{dalle}, Stable-diffusion \cite{stable_diffusion}, Unidiffuser \cite{bao2023one}, \textit{etc.}
However, due to the limitations of model capacity, computing overhead, and human-AI interaction gap, \textit{etc.,} not all generated images can satisfy users' requirements, and many AIGIs do not conform to human visual preferences or expectations.
Unlike common image content, such as Natural Scene Images (NSIs) \cite{9082850,min2024exploring}, Screen Content Images (SCIs) \cite{7444164,min2017unified} \textit{etc.}, which generally encounters common degradations such as noise, blur, compression \cite{gu2016blind,2024freqfusion,9271914,duan2022develop,2021coded,fu2023category,2023instance,min2018objective,2024hyperspectral,min2019quality}, AIGIs exhibit unique distortions including unreal structures, unreasonable component combinations, \textit{etc.} \cite{database/align:HPS,yang2024aigcoiqa2024,ding2020image}
Moreover, for text-to-image generation methods, the generated images may not accurately correspond to the semantics of the text prompts or may not meet the expectations of users \cite{zhou2024adaptive}.
\begin{figure}[t]

	\centering
	\includegraphics[width=\linewidth]{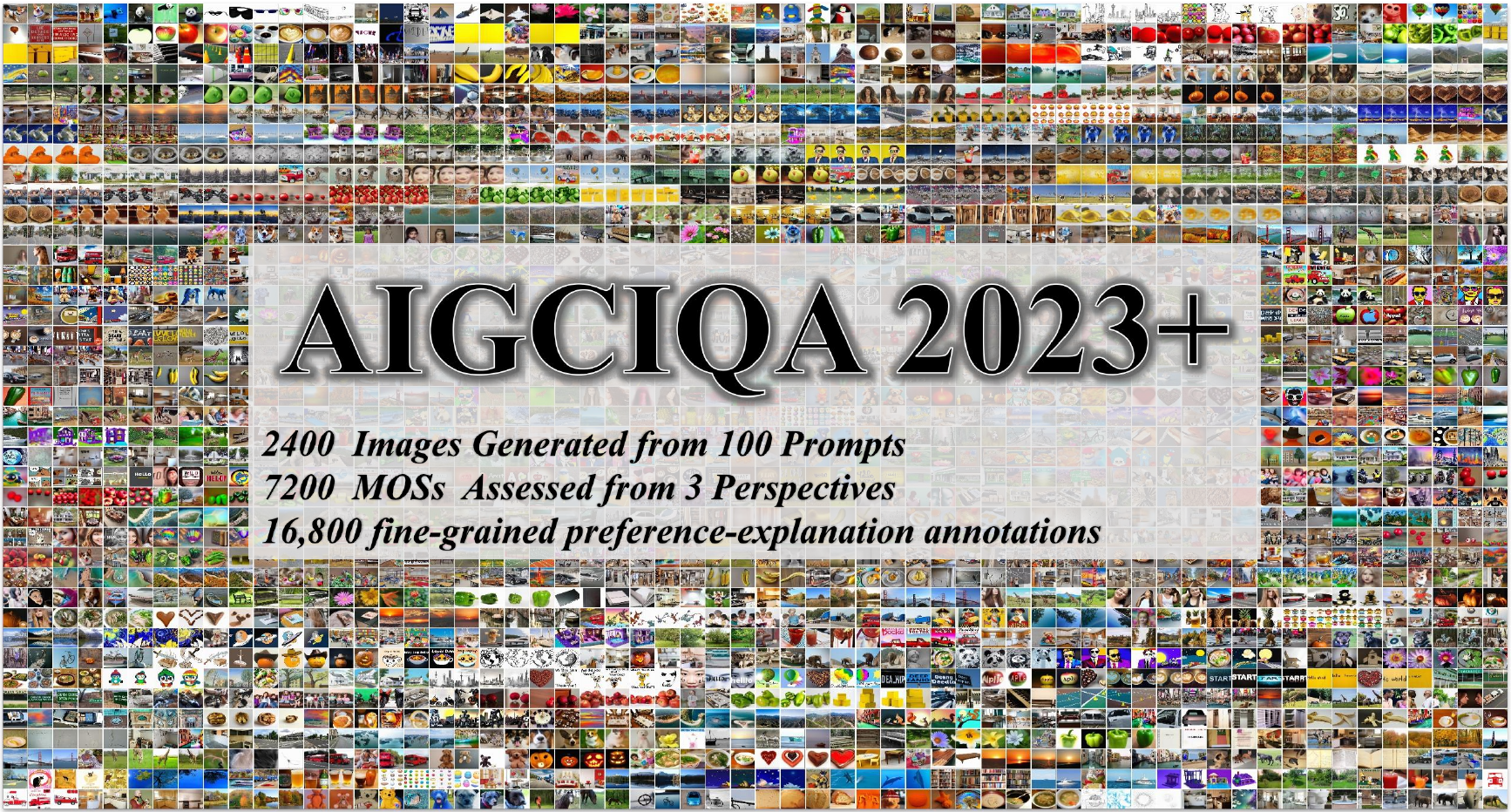}
         \vspace{-7mm}
	\caption{We present AIGCIQA2023+, a large-scale dataset that includes 2400 images generated from 100 prompts, 7200 MOSs from 3 perspectives, and 16,800 fine-grained preference-explanation annotations.}
	\label{AIGCIQA2023}
 \vspace{-3mm}
  
\end{figure}

Given these challenges, there is an urgent need to study and model human visual preferences and expectations for AIGIs, which can be used to monitor the quality of generated images, filter unexpected images, or even feedback to refine and improve the generation capability of AI models \cite{database/align:PickAPic}.
To this end, several Image Quality Assessment (IQA) databases for AIGIs such as ImageReward \cite{database/align:ImageReward}, AGIQA-3K \cite{database/agiqa}, AIGCIQA2023 \cite{wang2023aigciqa2023}, \emph{etc.}, have been proposed to characterize the human visual preferences for AIGIs by scoring or ranking these images. However, these works lack fine-grained preference explanations for AIGIs, which makes it hard to understand corresponding human visual preferences and carry out adaptive model/data adjustments for the generation methods.

Furthermore, several evaluation metrics have been proposed for the objective evaluation of AIGIs.
Inception Score (IS) \cite{gulrajani2017improved}, Fréchet Inception Distance (FID) \cite{heusel2017gans}, \textit{etc.}, have been widely used to evaluate the authenticity of the generated images.
However, these methods can not evaluate the authenticity of an individual generated image, and besides the authenticity properties, they cannot measure other human preference attributes such as quality and text-image correspondence.
Additionally, Language-image pre-training techniques such as CLIP \cite{radford2021learning}, BLIP \cite{li2022blip}, \textit{etc.}, have been widely used to evaluate the text-image correspondence recently, which ignores the visual experience attribute of AIGIs. Therefore, it is necessary to design a more comprehensive AIGI preference assessment model, which can evaluate both visual experience and text-image correspondence.

\begin{figure*}[t]
\vspace{-6mm}
	\centering
	\includegraphics[width=\linewidth]{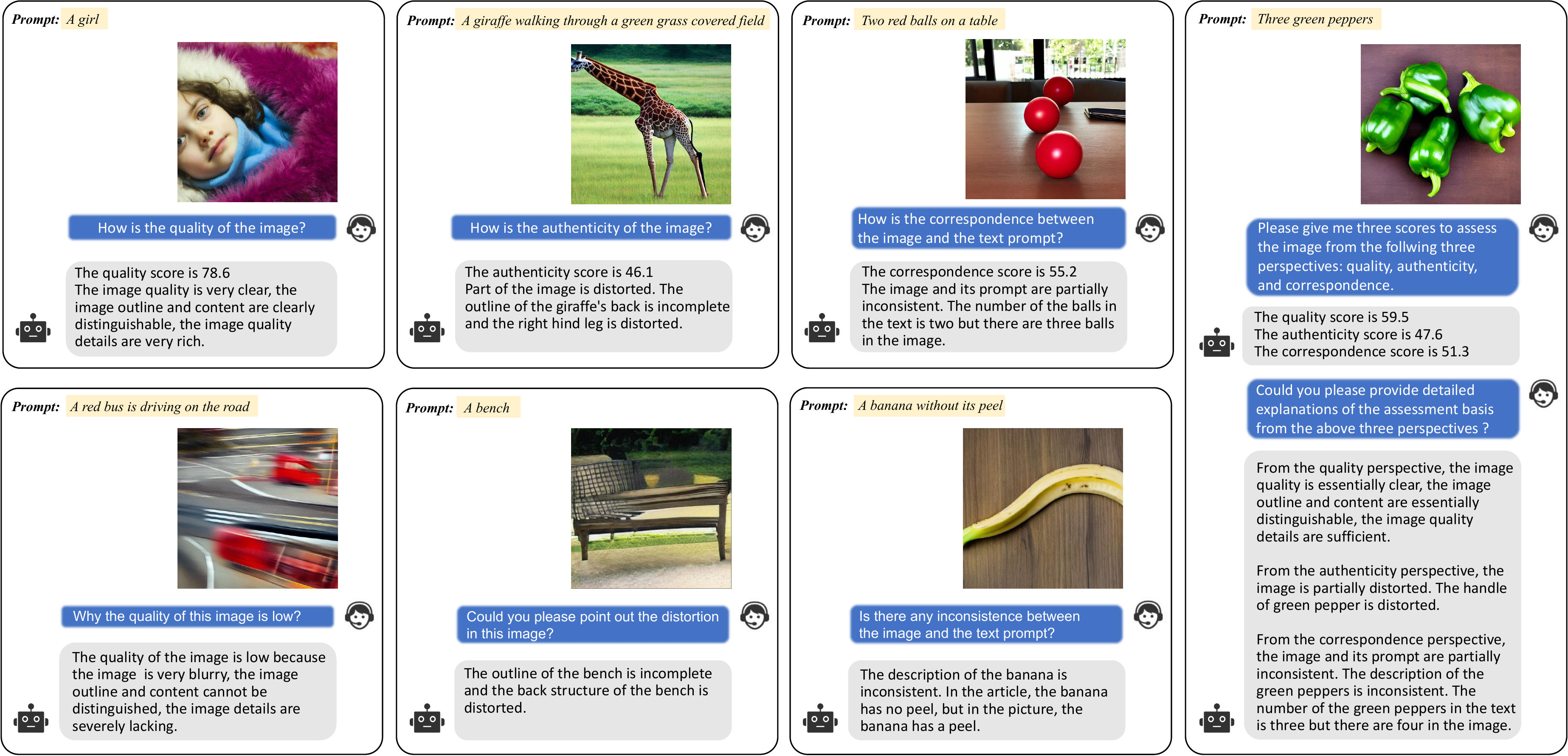}
 \vspace{-5mm}
	\caption{Examples generated by our MINT-IQA model, demonstrating its diverse capabilities including image quality assessment from multiple perspectives, abundant degradation-aware visual question answering, comprehensive human visual preference explaining, \textit{etc.}}
 \vspace{-3mm}
	\label{dialog}
\end{figure*}

To address the aforementioned challenges, in this paper, we first extend the previous AIGCIQA2023 database \cite{wang2023aigciqa2023} to AIGCIQA2023+, towards better understanding and evaluating human preferences for AIGIs, and then propose a novel method, termed \textbf{MINT-IQA}, towards automatically evaluating and explaining the visual preference and expectation attributes for AIGIs from \underline{M}ulti-perspectives with \underline{IN}struction \underline{T}uning. As an extension version of the conference version \cite{wang2023aigciqa2023}, the differences rely on two aspects: (1) annotation extension and (2) a new method design.
Specifically, as shown in Fig. \ref{AIGCIQA2023}, the established AIGCIQA2023+ contains 2400 AIGIs generated using 6 AIGI models based on 100 text prompts, 7,200 Mean Opinion Scores (MOSs) and 16,800 detailed preference-explanation annotations for these AIGIs.
The evaluation perspectives include quality, authenticity, and text-image correspondence.
We further propose a MINT-IQA model, which allows for a more comprehensive evaluation of the human preferences for AIGIs from multi-perspectives, resulting in state-of-the-art performance on AIGC IQA databases. 
% , not only limited to AI-generated images but also applicable for natural images.
To further enhance the degradation understanding and explanation capabilities of our model, we adopt the vision-language instruction tuning strategy, and fine-tune the MINT-IQA model based on the AIGCIQA2023+ database. This approach effectively leverages the frozen Large Language Model (LLM) to provide detailed explanations of the human preference basis for each AIGI as shown in Fig. \ref{dialog}, which can be used for feedback to further improve the assessment capabilities as shown in Section V. Additionally, MINT-IQA model also achieves the best performance on traditional IQA databases, which further demonstrates its generality in understanding and evaluating human preferences.

%adding a layer of interpretability to the evaluation process.

The main contributions of this paper can be summarized as follows:
\begin{itemize}
\item We establish an AI-generated image quality assessment database, termed AIGCIQA2023+, to conduct comprehensive understanding and evaluation research on human visual preferences for AIGIs.
% \item We propose to disentangle the human visual experience for AIGIs into three perspectives including quality, authenticity, and correspondence. 
% Based on the above theory, we establish a novel database, i.e., MINT-IQA, to better understand the human IQA basis for AIGIs and instruct the LLM for explanation.
\item We propose a MINT-IQA model, which integrates text and image information via Q-Former to evaluate the human visual preference from multiple perspectives including quality, authenticity, and text-image correspondence.

\item We utilize the vision-language instruction tuning strategy to enable MINT-IQA to better understand and explain human visual preference on AIGIs, which can be used to further improve the evaluation ability.

\item Extensive experimental results demonstrate that our proposed MINT-IQA model achieves state-of-the-art performance on both AIGC IQA databases and traditional IQA databases, and attains comprehensive understanding and explanation capabilities for human visual preference of AIGIs.
\end{itemize}
\vspace{-5mm}
% \vspace{-3mm}
\section{Related Work}
\vspace{-1mm}
\label{sec:formatting}
% The related work is divided into two parts according to the contributions of the article. 
% After briefly introducing the IQA databases used in our study, we review state-of-the-art blind IQA methods for comparison.
\subsection{Image Quality Assessment Databases}
\vspace{-1mm}
\begin{table*}[tbph]
\vspace{-4mm}
		\centering
  \footnotesize
		%Publicly Available
		\caption{Comparison of image quality assessment databases, including traditional IQA databases and AIGC IQA databases.}
		\label{tab:database}
  \vspace{-2mm}
  \scalebox{1.1}{
		\begin{tabular}{c|c|c|c|c|c|c}
			\toprule
			Type & Database & Score              & Preference annotation             & Image   & Ratings   & Explanations                       \\ \hline
			% DiffsionDB \cite{database:DiffusionDB}  & No                 & No                    & 1,819,808 & 0       & Diffusion (1)                             \\ \hline
   %              HPS \cite{database/align:HPS}        & Preference         & Overall               & 98,807   & 98,807   & Diffusion (1)                            \\ \hline
			% Pick-A-Pic \cite{database/align:PickAPic}  & Preference         & Overall               & 500,000  & 500,000  & Diffusion (3)                             \\ \hline
			
				\multirow{8}{*}{\begin{tabular}[c]{@{}c@{}}Traditional\\IQA Databases\end{tabular}} 
                &LIVE \cite{sheikh2006statistical}& MOS & Overall&779& 25,000 & No\\\cline{2-7}
                &CSIQ \cite{larson2010most}& MOS & Overall&866& 5,000& No\\\cline{2-7}
                &TID2013 \cite{ponomarenko2015image}& MOS & Overall&3,000& 524,340 & No\\\cline{2-7}
                
                &CLIVE \cite{ghadiyaram2015massive}& MOS & Overall&1,162& 350,000& No\\\cline{2-7}
                &KADID \cite{lin2019kadid}& MOS & Overall&10,125& 303,750& No\\\cline{2-7}
                % &CID2013 \cite{virtanen2014cid2013}& Rating & Overall&480&14,880& No\\\cline{2-7} 
               &KonIQ-10k \cite{hosu2020koniq} & MOS &Overall& 10,073 & 1,208,760  & No\\\cline{2-7} 
                &SPAQ \cite{fang2020perceptual} & MOS & Overall & 11,125 & 55,625  & No\\\cline{2-7} 
                &AVA \cite{ava} & Rating & Overall & 255,000 & 53,550,000 & No \\ \hline
                \multirow{6}{*}{\begin{tabular}[c]{@{}c@{}}AIGC\\IQA Databases\end{tabular}} &
                DiffsionDB \cite{database:DiffusionDB}  & No                 & No                    & 1,819,808 & 0                            & No\\  \cline{2-7}
               & HPS \cite{database/align:HPS}        & Preference         & Overall               & 98,807   & 98,807                       & No\\\cline{2-7}
			& Pick-A-Pic \cite{database/align:PickAPic}  & Preference         & Overall               & 500,000  & 500,000              & No  \\ \cline{2-7}
                &ImageReward \cite{database/align:ImageReward} & Ranking & Overall & 136,892  & 410,676           & No \\\cline{2-7}
				&AGIQA-3K \cite{database/agiqa}    & MOS                & Perception, alignment & 2,982    & 125,244  & No\\ \cline{2-7}
                
                &AIGCIQA2023 \cite{wang2023aigciqa2023}    & MOS                & Quality, authenticity, correspondence & 2,400    & 201,600     & No\\  \cline{2-7}
                &AIGCIQA2023+    & MOS                & Quality, authenticity, correspondence & 2,400    & 201,600    &  16,800\\  
               
                \bottomrule
			\end{tabular}}
    \vspace{-4mm}
		\end{table*}

Many IQA databases have been established in the literature  \cite{duan2023attentive,duan2024quick,min2024perceptual,zhai2020perceptual,min2020study}. As shown in Table \ref{tab:database}, according to the image acquisition method, they can be categorized into two types: (1) traditional IQA databases and (2) AIGC IQA databases.
Traditional IQA databases consist of natural images with realistic camera distortions such as noise, blur, compression, \textit{etc.} 
% CID2013 \cite{virtanen2014cid2013} is a small database consists of 585 realistically blurred pictures taken by 79 cameras.  
LIVE \cite{sheikh2006statistical}, CSIQ \cite{larson2010most}, TID2013 \cite{ponomarenko2015image}, and CLIVE \cite{ghadiyaram2015massive} are fundamental traditional IQA databases and have been widely used for developing IQA models.
 KADID \cite{lin2019kadid}, KonIQ-10k \cite{kong2016photo} and SPAQ \cite{fang2020perceptual} expand the data scale and cover a wide range of scene categories. AVA \cite{ava} is a large-scale database constructed for aesthetic image quality assessment. Recently, the development of text-to-image generative models has led to an explosion of AI-generated images.
Thus many AIGC IQA databases have also been constructed, which contain generated images with unique distortions including unreal structures, unreasonable component combinations, \textit{etc.} DiffusionDB \cite{database:DiffusionDB} is a large-scale database containing over 1.8 million text-image pairs without score-specific annotations.
Pick-A-Pic \cite{database/align:PickAPic} and HPS \cite{database/align:HPS} provide subjective annotations of human preferences by selecting the most preferred one among a group or a pair of AIGIs.
ImageReward \cite{database/align:ImageReward} contains preference ranking annotations for AIGIs.
The aforementioned AIGC IQA databases only evaluate the AIGIs from the overall dimension, which cannot reflect and characterize complex human visual preferences for AIGIs.
% In addition to diffusion-based models, it also includes auto-regression-based models. 
AGIQA-3K \cite{database/agiqa} refines the rating granularity by scoring two dimensions including perception and alignment. Our preliminary AIGCIQA2023 database \cite{wang2023aigciqa2023} contains subjective quality ratings from three dimensions including quality, authenticity, and correspondence.
However, the aforementioned works lack detailed annotations of the human preference for AIGIs, which motivates us to further extend the AIGCIQA2023 database to include language annotations.
% Consequently, a fine-grained AIGC IQA database including both accurate image quality scores from multiple perspectives and detailed explanations of the quality factors that influence human preference is needed. 
% The above issue motivates us to construct a new database for AIGIs collection, evaluation, and explanation, which aims to cover more accurate human preference scores from different perspectives and provide more detailed explanations to guide the design process of objective IQA models, which needs the capabilities of giving accurate scores and explaining why this score is given.
\vspace{-3mm}

\subsection{Image Quality Assessment Models}
\vspace{-1mm}
Numerous image and video quality assessment models have been proposed in the literature \cite{fu2024vision,guan2023dual,guan2022visibility,wu2023q,tu2021ugc,wu2022fast,tu2021rapique,zheng2024faver,9921340,zheng2024video,min2018blind,duan2024finevq,wang2024aigv,duan2022confusing,xu2025harmonyiqa}, which aims to objectively and quantitatively evaluate the perceptual quality.
% Image quality assessment algorithms aim to objectively and quantitatively evaluate the perceptual quality of images. 
Based on different feature extraction and evaluation methods, current popular models for image quality assessment can be classified into three categories: (1) Handcrafted-based models, including: NIQE \cite{mittal2012making}, BRISQUE \cite{mittal2012no}, QAC \cite{xue2013learning}, BMPRI \cite{quality:BMPRI}, HOSA \cite{xu2016blind}, BPRI \cite{min2017blind},    HIGRADE \cite{kundu2017large}, \textit{etc.} These models generally extract features based on prior knowledge related to image quality. (2) Deep learning-based models, including: CNNIQA \cite{kang2014convolutional}, DBCNN \cite{quality:DBCNN}, WaDIQaM-NR \cite{bosse2017deep}, StairIQA \cite{sun2023blind}, CLIPAGIQA \cite{fu2024vision}, Q-Bench~\cite{wu2023q}, \textit{etc.} These models characterize quality-aware information by training deep neural networks using labeled data. (3) With the popularity of text-to-image generation, many vision-language pre-training models have been adopted to evaluate the text-image alignment, including: CLIP \cite{radford2021learning}, BLIP \cite{li2022blip}, FLIP \cite{li2023scaling}, Aesthetic \cite{schuhmann2022laion}, HPS \cite{database/align:HPS}, PickScore \cite{database/align:PickAPic}, ImageReward \cite{database/align:ImageReward}, StairReward \cite{database/agiqa}, \textit{etc.} These models are often used for evaluating generated images based on its text prompt considering a mixture of elements such as text-image alignment, fidelity, aesthetics, \textit{etc.} 
% In addition to IQA, Video Quality Assessment (VQA) algorithms have been developed to evaluate the perceptual quality of video content. Notable VQA models include UGC-VQA~\cite{tu2021ugc}, FAST-VQA~\cite{wu2022fast}, RAPIQUE~\cite{tu2021rapique},  FAVER~\cite{zheng2024faver}, and VIQE \cite{9921340}. These VQA models address various challenges such as handling user-generated content, ensuring computational efficiency, and providing accurate predictions under diverse conditions.
However, the aforementioned IQA models can only evaluate a single-dimension preference for AI-generated images or videos, which lacks the ability to conduct a comprehensive assessment from multiple perspectives.  This limitation motivates us to propose a unified model to evaluate AIGIs from multiple dimensions.

% The aforementioned models can only evaluate a single-dimension preference for AIGIs, which lacks the ability to conduct a comprehensive assessment from multiple perspectives. 
% This motivates us to propose a unified model to evaluate AIGIs from multiple dimensions.
  % hand-crafted feature based models, language-Image pre-training models and deep learning based models.
% \vspace{-3mm}
% \input{figures/sample}

\section{Database Construction and Analysis}
% \vspace{-2mm}
\begin{figure}[t]
% \vspace{-5mm}
	\centering
	\includegraphics[width=\linewidth]{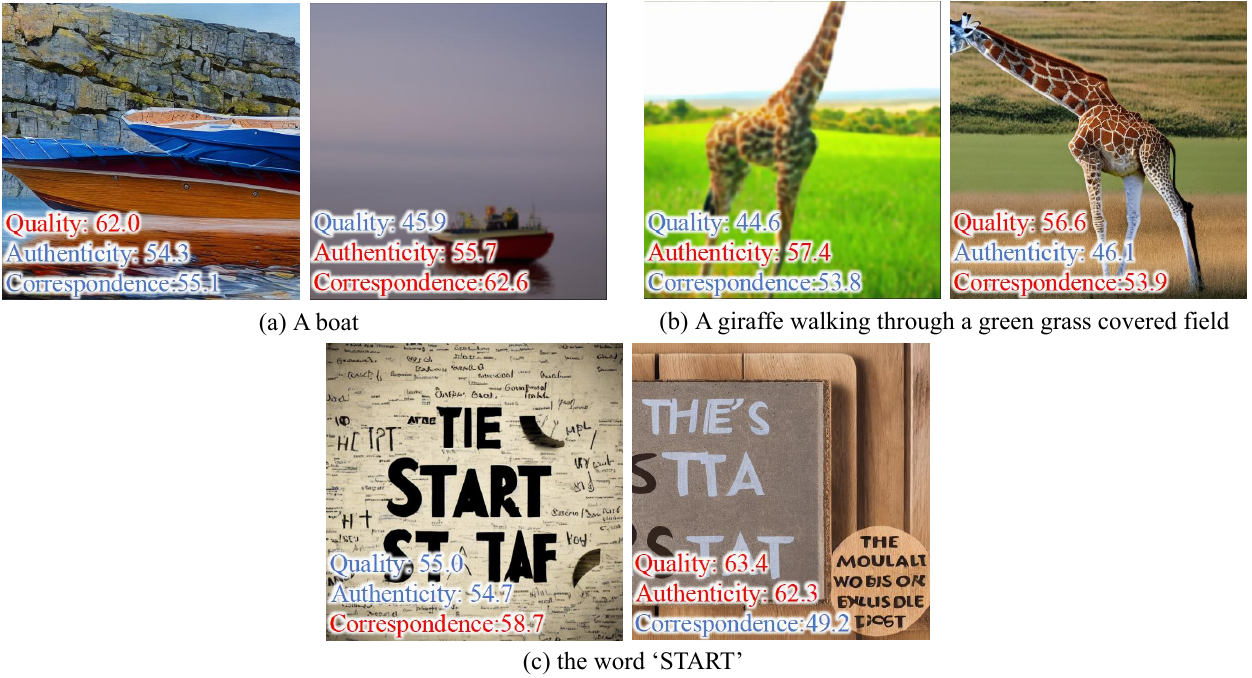}
        \vspace{-7mm}
	\caption{Rating comparisons between three perspectives. (a) The quality score of the left image is higher, but other two scores are lower. (b) The authenticity score of the left image is higher, but other two scores are lower. (c) The correspondence score of the left image is higher, but other two scores are lower. }
	\label{comparer}
 \vspace{-3mm}
  
\end{figure}
This work focuses on understanding and evaluating human preferences for AI-generated images, therefore it is necessary to construct a database that provides various AI-generated images with accurate human preference scores and detailed explanations for evaluation factors. In order to better understand human preferences for AIGIs, we propose to disentangle the human visual experience of AIGIs into three perspectives, including quality, authenticity, and correspondence, and then conduct scoring and interpretation. “Quality” evaluates an AIGI from the visual quality attribute considering the brightness, colorfulness, clarity, \textit{etc.}, “authenticity” evaluates the reality-degree attribute of an AIGI including naturalness, harmonious, \textit{etc.}, “correspondence” assesses the relevance attribute between an AIGI and its prompt. As shown in Fig. \ref{comparer}, different dimensions can reflect different human preference characteristics, which further strength the importance of evaluating AIGIs from multiple perspectives. Overall, our constructed AIGCIQA2023+ dataset contains 2,400 AI-generated images with 7,200 corresponding MOSs evaluated from three perspectives and 16,800 detailed interpretations of evaluation criteria.
Detailed subjective experimental procedures are described as follows.
\vspace{-3mm}
\label{AIGI Collection}
\begin{figure*}[!t]
\vspace{-4mm}
	\centering
	\includegraphics[width=\linewidth]{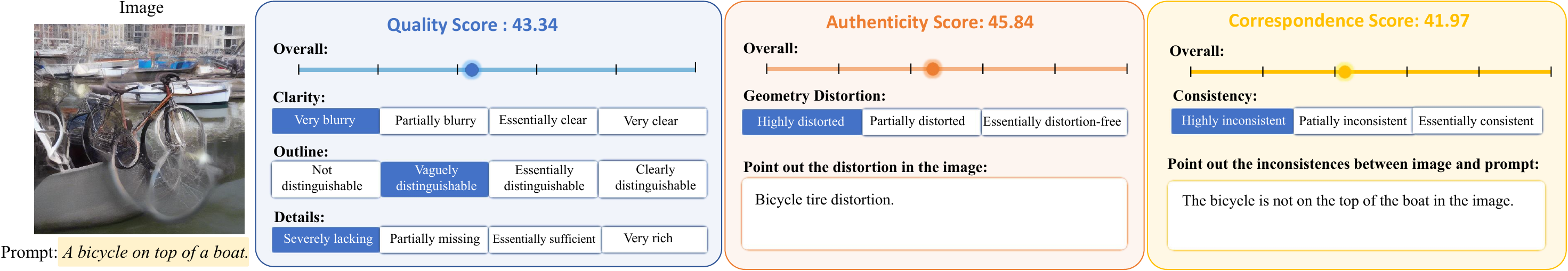}
 \vspace{-7mm}
	\caption{Illustration of the subjective assessment interface. The subjects are instructed to make fine-grained assessments and annotations by clicking the checkboxes and give further detailed explanations by inputting text descriptions.}
 \vspace{-3mm}
	\label{ui}
\end{figure*}
\subsection{AIGI Collection}

We adopt six latest text-to-image generative models, including Glide\cite{Nichol2021GLIDETP}, Lafite\cite{Zhou_2022_CVPR}, DALLE\cite{ramesh2022hierarchical}, Stable-diffusion\cite{Rombach2021HighResolutionIS}, Unidiffuser\cite{bao2023one}, Controlnet\cite{Zhang2023AddingCC}, to produce AIGIs by using open source code and default weights.
To ensure content diversity and catch up with the practical application requirements, we collect diverse texts from the PartiPrompts website \cite{yu2022scaling} as the prompts for AI-based image generation.
The text prompts can be simple, allowing generative models to produce imaginative results.
They can also be complex, which raises the challenge for generative models.
We select 10 scene categories from the prompt set, including people, animals, artifacts, illustrations, indoor scenes, outdoor scenes, vehicles, produce \& plants, food \& beverage, and world knowledge. Each scene contains 10 challenge categories, including basic, simple detail, fine-grained detail, complex, quantity, imagination, style \& format, perspective, writing \& symbols, and linguistic structures.
Overall, we collect 100 text prompts (10 scene categories $\times$ 10 challenge categories) from PartiPrompts\cite{yu2022scaling}.
% The distribution of the selected scene and challenge categories is displayed in pie chart of Fig.1.
% It can be observed that the dataset exhibits a high level of scene diversity, with images generated covering a broad range of challenges.
% Then we perform the text-to-image generation based on these models and prompts. 
Based on the selected prompts, we generate 4 different images for each generative model. Therefore, the constructed database totally contains 2400 AIGIs (4 images $\times$ 6 models $\times$ 100 prompts).
\vspace{-3mm}
\subsection{Subjective Experiment Setup}

% \vspace{-2mm}
Based on the collected AIGIs, we further conduct a subjective experiment following the guidelines of ITU-R BT.500-14 \cite{duan2022confusing}, to obtain subjective quality ratings from the perspectives of quality, authenticity, and text-image correspondence. 
The detailed explanations for the three perspectives can be found above in Section \ref{AIGI Collection}.
Moreover, to obtain fine-grained preference-related annotations, subjects are instructed to give detailed explanations by finishing the fill-in-the-blank questions.
Specifically, for the quality perspective, the subjects are asked to give detailed description choices for the degree of image clarity,  the outline and content of the image, and the richness of details. As for the authenticity perspective, subjects are instructed to assess the distortion severity and point out the specific distorted part of each AIGI. To evaluate the text-image correspondence, subjects are instructed to consider whether there is any inconsistency between the image content and its text prompt and then input detailed explanations through typing. In our subjective experiment, there are five single-choice questions designed. Each question corresponds to a factor that influences human preference for each AIGI. We provide several choices to define the level of each factor. For image clarity, we divide it into four levels, including: very blurry, partially blurry, essentially clear, and very clear. Similarly, for the factors of the image outline and details, we also divide them into four levels.
% , as shown in Figure \ref{bus}. 
Furthermore, the geometry distortion and the text-image consistency are divided into three levels, including: highly (distorted / inconsistent), partially (distorted / inconsistent), and essentially (distortion-free / consistent).

The AIGIs are presented in a random order on an iMac monitor with a resolution of 4096 × 2304, using an interface designed with Python Tkinter as shown in Fig. \ref{ui}. The interface allows viewers to browse the previous and next AIGIs, and evaluate them by scoring within the range from 0 to 5 as well as providing fine-grained explanations. A total of 76 graduate students participate in the experiment. The subjects are seated at a distance of around 60 cm in a laboratory environment with normal indoor lighting. 
To ensure consistency, we have developed and documented the annotation process. For the purpose of explaining human preferences, we have also established criteria for providing explanations in a standardized format, and have included numerous illustrative examples. 
The subjects are provided with detailed evaluation criteria and trained before the formal experiment to be familiar with the experiment.
Individuals who demonstrate low agreement accuracy are excluded from the experiment. Finally, we collect 84 (28 subjects × 3 perspectives) ratings and an average of 7 preference descriptions for each AIGI, resulting in a total number of 201,600 (28 × 3 × 2400) quality ratings and 16,800 (7 × 2,400) detailed explanations.%by inputting text descriptions. 
% Since an AIGI is generated from a given text prompt, it is also important 
% To evaluate the text-image correspondence, subjects are instructed to consider whether there is any inconsistency between the image content and its text prompt and then input detailed explanations through typing.

\vspace{-3mm}

\subsection{Subjective Rating Processing}

We follow the suggestions recommended by ITU to conduct the outlier detection and subject rejection. 
The score rejection rate is 2\%.
In order to obtain the MOS for an AIGI, we first convert the raw ratings into Z-scores, then linearly scale them to the range $[0,100]$ as follows:
\begin{eqnarray}
\begin{aligned}
z_i{}_j=\frac{r_i{}_j-\mu_j}{\sigma_i}, ~\quad z_{ij}'=\frac{100(z_{ij}+3)}{6},
\end{aligned}
\end{eqnarray}
\begin{eqnarray}
\begin{aligned}
\mu_j=\frac{1}{N_i}\sum_{i=1}^{N_i}r_i{}_j, ~\sigma_j=\sqrt{\frac{1}{N_i-1}\sum_{i=1}^{N_i}{(r_i{}_j-\mu_j)^2}},
\end{aligned}
\end{eqnarray}
where $r_{ij}$ is the raw ratings given by the $i$-th subject to the $j$-th image. $N_i$ is the number of images judged by subject $i$. Next, the mean opinion score (MOS) of the image j is computed by averaging the rescaled z-scores as follows:
\begin{eqnarray}
\begin{aligned}
MOS_j=\frac{1}{M}\sum_{i=1}^{M}z_{ij}',
\end{aligned}
\end{eqnarray}
where $MOS_j$ indicates the MOS for the $j$-th AIGI, $M$ is the number of valid subjects, and $z'_i{}_j$ are the rescaled z-scores.
% \vspace{-2mm}
% The AIGIs are presented in a random order on a monitor with a resolution of up to 4096 × 2304, using an interface designed with Python Tkinter, as shown in \textit{supplementary material (Supp.)}. A total of 48 graduate students participate in the experiment. The subjects are trained before experiment and provided with detailed evaluation criteria. Finally, we collect an average of 7 pieces of information related to human preference for each AIGI, resulting in a total number of 16,800 (7 × 2,400) detailed explanations of the human preferences from three different perspectives.
\vspace{-3mm}

\begin{figure*}[t]
% % \vspace{-5mm}
	\centering
	\includegraphics[width=\linewidth]{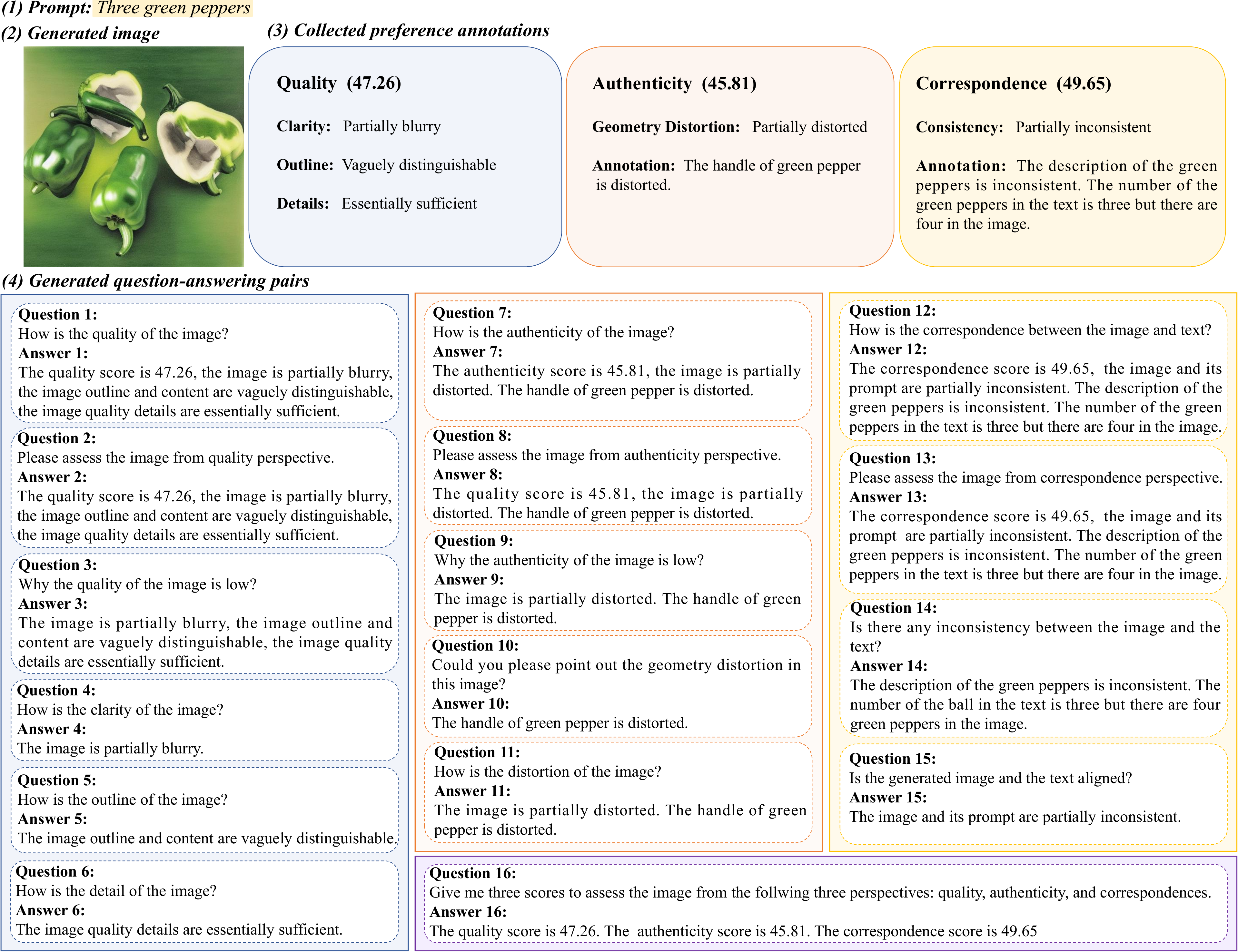}
\vspace{-7mm}
	\caption{Illustration of the generated visual question-answering pairs for instruction tuning.}
	\label{VQA}
 % \vspace{-2mm}
\end{figure*}

\subsection{Question Answering Pair Generation}
% \vspace{-2mm}

In order to enable the detailed human preference explanation ability for an IQA model, we collect numerous question-answering pairs tailored for instruction tuning.
% The answers related to the detailed explanations 
Some of the questions are consistent with the setup questions in our subjective experiment, and some are extended based on the degradation or quality-related labels. The extended question-answering pairs are generated with the help of LLM (LLaMA\cite{touvron2023llama}).
Specifically, we feed the collected quality labels and detailed explanations into LLM, then prompt LLM to ask questions for these answers.
This step can greatly expand the scale of VQA pairs.
Fig. \ref{VQA} fine-grained human preference annotations, we can improve the perceptual explanation capability of LLMs and further improve the performance of IQA models.

%Please refer to \textit{the supplementary material} for more details.
% detailed explanations to questions come from the database we built. Considering the sensitivity of large language models to symbolic input and the differences in individual answers, we use standardized templates to unify the format of the collected questions and answers related to image quality factors.
% we generate various quality related questions and answers for training the model. To avoid the risk of model overfitting, we follow the InstructBLIP to craft 10 to 15 distinct instruction templates in natural language to articulate the task and the objective.

% For instruction tuning, we devise numerous question-answering pairs by matching the instruction questions to the corresponding answers.
% Considering the sensitivity of LLM to the labeling formats, we craft several standardized templates to unify the format of the collected questions and their corresponding answers and organize them into pairs. 
% Some examples are illustrated in Figure \ref{VQA}.
% By leveraging the insights and knowledge gained from our AIGCIQA2023 database and organize them into standard question-answering pairs for instruction tuning, we can enhance the accuracy and reliability of IQA models in predicting and explaining human visual preferences. 

% \input{figure/bus}

% \section{Human Annotation Analysis}
\vspace{-3mm}
\subsection{Human Annotation Analysis}
\begin{figure*}[t]
% \vspace{-4mm}
%   \centering
%   \includegraphics[width=\textwidth]{figures/s1.png}
%   \caption{ Scores and MOSs probability distribution of quality score.}
% ~\\
% ~\\
%   \centering
%   \includegraphics[width=\textwidth]{figures/s2.png}
%   \caption{ Scores and MOSs probability distribution of reality score.}
% ~\\
% ~\\
%   \centering
%   \includegraphics[width=\textwidth]{figures/s3.png}
%   \caption{ Scores and MOSs probability distribution of relation score.}
  \centering
  \includegraphics[width=\linewidth]{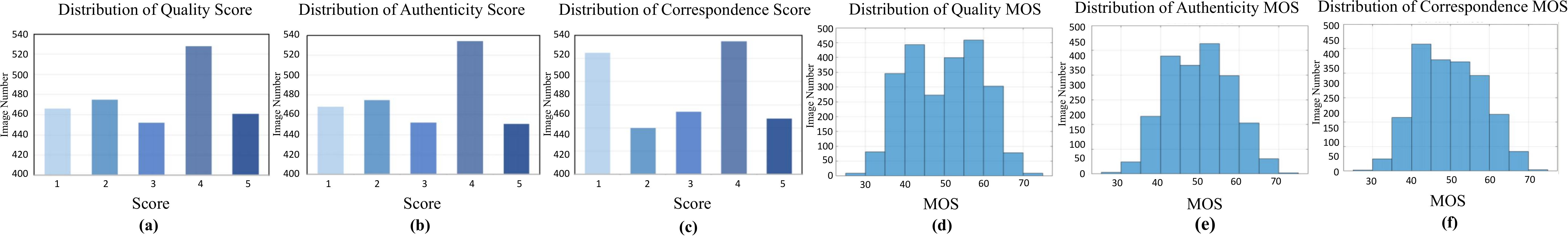}
    \vspace{-8mm}
  \caption{ (a) Distribution of the quality scores.
  (b) Distribution of the authenticity scores.
  (c) Distribution of the correspondence scores. 
  (d) Distribution of the quality MOSs.
  (e) Distribution of the authenticity MOSs.
  (f) Distribution of the text-image correspondence MOSs.
  }
  \vspace{-1mm}
  \label{scores}
\end{figure*}

\begin{figure*}[t]
% \vspace{-5mm}
	\centering
	\includegraphics[width=\linewidth]{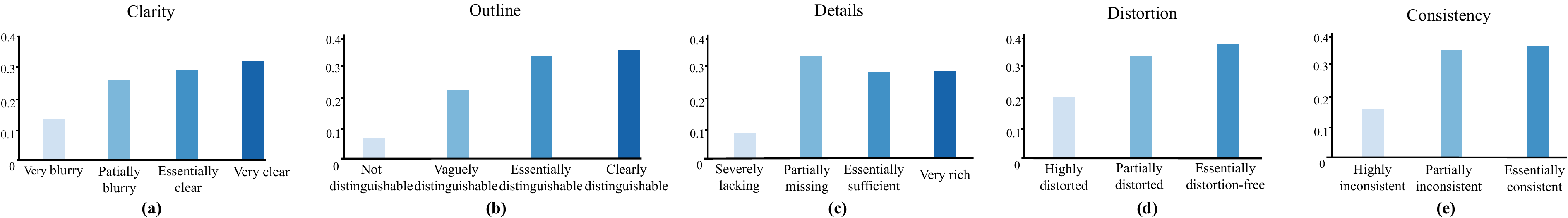}
 \vspace{-8mm}
	\caption{Distribution of fine-grained annotations. (a) Distribution of the clarity. (b) Distribution of the outline recognizability. (c) Distribution of the richness of details. (d) Distribution of the geometry distortion. (e) Distribution of the text-image consistency. }
 \vspace{-1mm}
	\label{distribution}
\end{figure*}

\begin{figure*}[t]
% \vspace{-6mm}
	\centering
	\includegraphics[width=\linewidth]{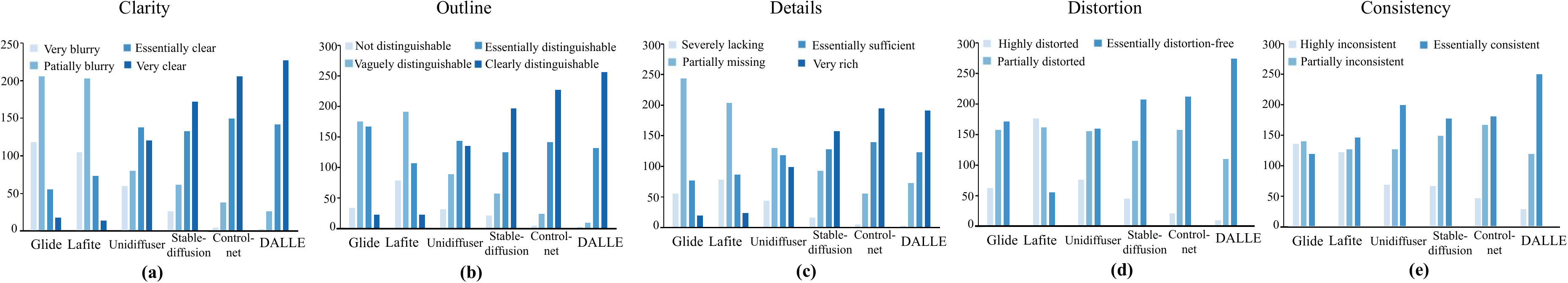}
 \vspace{-8mm}
	\caption{Distribution of fine-grained annotations for different generation models including Glide \cite{Nichol2021GLIDETP}, Lafite \cite{Zhou_2022_CVPR}, Unidiffuser \cite{bao2023one}, Stable-diffusion \cite{Rombach2021HighResolutionIS}, Controlnet \cite{Zhang2023AddingCC}, DALLE \cite{ramesh2022hierarchical}. (a) Distribution of the clarity. (b) Distribution of the outline recognizability. (c) Distribution of the richness of details. (d) Distribution of the geometry distortion. (e) Distribution of the text-image consistency. }
 % \vspace{-2mm}
	\label{dc}
\end{figure*}

% \vspace{-5mm}
We first analyze the subjective quality rating and MOS distribution from the perspectives of quality, authenticity, text-image correspondence respectively.
As shown in Fig. \ref{scores}, the images in AIGCIQA2023+ database cover a wide range of perceptual quality. 
% In our subjective experiment, there are five single-choice questions designed. Each question corresponds to a factor that influences human preference for each AIGI. We provide several choices to define the level of each factor. For image clarity, we divide it into four levels, including: very blurry, partially blurry, essentially clear, and very clear. Similarly, for the factors of the image outline and details, we also divide them into four levels.
% % , as shown in Figure \ref{bus}. 
% Furthermore, the distortion and the text-image consistency are divided into three levels, including: highly (distorted / inconsistent), partially (distorted / inconsistent), and essentially (distortion-free / consistent).
We further analyze the distributions of fine-grained annotations. Fig. \ref{distribution} demonstrates the distribution of fine-grained annotations for all models, and Fig. \ref{dc} further shows the distribution of fine-grained annotations for different generation models. It can be observed that the performance of DALLE \cite{ramesh2022hierarchical} is the best among all the six AIGI models. The performance of Glide \cite{Nichol2021GLIDETP} and Lafite \cite{Zhou_2022_CVPR} is relatively poor compared to other models in terms of the fine-grained assessment. Besides,
Lafite \cite{Zhou_2022_CVPR} is better than Glide \cite{Nichol2021GLIDETP} in terms of text-image consistency, but far worse than Glide \cite{Nichol2021GLIDETP} in terms of the authenticity perspective. This also manifests that different AIGI models have different defects, which further illustrates the importance of evaluating AIGIs from multiple perspectives. And our analysis can help users choose appropriate AIGI models according to the different preference emphasis.
\vspace{-1mm}

\section{Proposed Method}
\begin{figure*}[t]
% \vspace{-4mm}
	\centering
	\includegraphics[width=\linewidth]{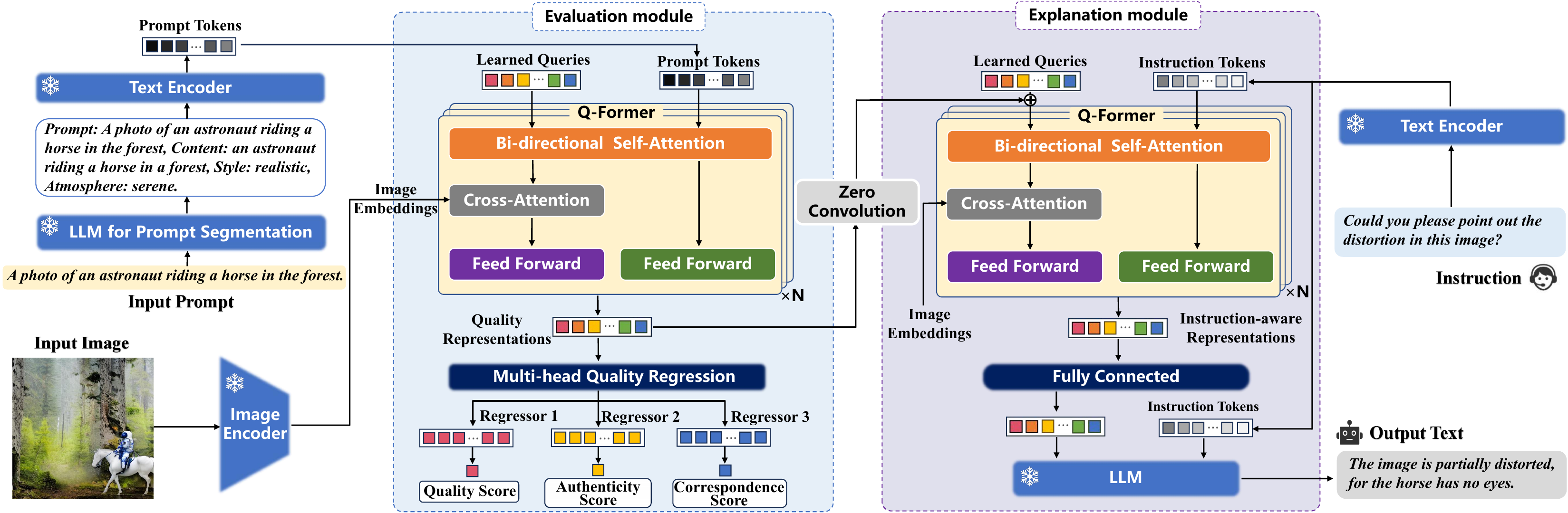}
	\caption{An overview of the architecture of the proposed MINT-IQA model. The MINT-IQA model consists of three parts: (1) For a given prompt and its generated image, we first use LLMs to segment the prompt to be more understandable, then use a text encoder and an image encoder to extract text and image features, respectively. (2) The extracted text and image features are interacted with each other through a cross-modal Q-Former to capture preference-related representations which are then separately regressed to various preference scores. (3) To enable the preference-related explanation capability for MINT-IQA, the extracted preference representations are interacted with the extracted instruction features via another Q-Former to obtain instruction-aware representations, which are then fed into a LLM to give answers and explanations.}
	\label{framework}
  % \vspace{-2mm}
\end{figure*}

We propose a MINT-IQA model to evaluate and explain human preferences from multiple perspectives with instruction tuing, which contains two main functions: (1) predicting human preference scores for AIGIs from multiple perspectives; (2) providing detailed explanations of the factors influencing human preference according to the questions. 
The network architecture of MINT-IQA is demonstrated in Fig. \ref{framework}. Specifically, the MINT-IQA model consists of three parts: (1) Given prompt and its generated image, we first use LLMs to segment the prompt to be more understandable, then encode the refined prompt and the generated image into text tokens and image embeddings using a text encoder and a image encoder, respectively. (2) For learning to evaluate the quality of the image, the extracted text and image features are interacted with each other through a cross-modal querying transformer (Q-Former) to capture preference representations which are separately mapped to various preference scores through different regressors. (3) To further understand and explain the evaluation process, another Q-Former is applied before a frozen LLM to perform instruction tuning and output answers for a given instruction question.
Detailed methods are given as follows.
\vspace{-3mm}
\subsection{Learning and Evaluating}
% For learning and evaluating, we use a LLM for prompt segmentation, a text encoder and an image encoder for textual and visual feature extraction, a Q-Former for multimodal representation learning, and three score regressors for mapping the interacted features to three different score spaces. These components enable the model to predict accurate human preference scores for each AIGI from multiple perspectives.
Given an AIGI and its corresponding prompt, we first segment the prompt to be more understandable and explicit through a LLM. Then we use a text encoder and an image encoder for textual and visual feature extraction. The extracted text and image features are interacted with each other through a cross-modal Querying Transformer (Q-Former) to learn multimodal quality representations which are then given to regressors to predict human preference scores from multiple perspectives.
% For learning and evaluating, we use a LLM for prompt segmentation, a text encoder and an image encoder for textual and visual feature extraction, a Q-Former for multimodal representation learning, and three score regressors for mapping the interacted features to three different score spaces. 
% Given an AIGI and its corresponding prompt, we first segment the prompt to be more understandable and explicit through a LLM. Then we use a text encoder and an image encoder for textual and visual feature extraction. The extracted text and image features are interacted with each other through a cross-modal Querying Transformer (Q-Former) to learn multimodal quality representations which are then given to regressors to predict human preference scores from multiple perspectives.
% The MINT-IQA framework, as depicted in Fig. \ref{framework}, consists of four parts: (1) a text encoder and an image encoder used to extract visual and textual features; (2) two cross-modal Querying TransFormers (Q-Formers) for feature interaction and image-text matching; (3) two Large Language Models (LLMs), one for prompt segmentation and another for vision-language generation; and (4) three score regressors for mapping the interacted features to three different score spaces.
% \subsubsection{Prompt Segmentation and Feature Extraction }
% why segmentation?

\subsubsection{Prompt Segmentation}
The content and length of the input prompt are of great significance in the text-image correspondence performance evaluation.
% Unlike descriptions obtained from captioning real images, text prompts for image generation sometimes tend to be highly detailed and specific, going beyond merely describing salient content.
The information provided by the original prompt may not be cohesive or meaningful enough, thus may lead to poor performance on the evaluation of AIGIs, particularly in terms of text-image correspondence. To address this issue, we apply a “raw prompt + LLM-refined prompt” mechanism to make the text more understandable for text models.
Specifically, we utilize a LLM (LLaMA \cite{touvron2023llama}) to transform text prompts into three task-specific annotations including “style”, “content”, and “atmosphere”.
“Style” words describe the whole image style, including painting, artistic, realistic, fashion, texture, fiction, \textit{etc.} If the original input text does not have style cues, the default style is “realistic”.
“Content” words are nouns with adjectives describing the image content. 
“Atmosphere” words refer to the emotional and psychological elements associated with the image, including mood and feeling conveyed by the scene. 
All of these three items lying after the raw prompt are crucial to the assessment of text-image correspondence.
% All of these three items lie beneath the prompt are crucial to the generation quality of the AIGI which is reflected in both human visual perception and text-image correspondence.
% Then we encode the text into text tokens and transform the input image into image embeddings.

\subsubsection{Feature Extraction} The feature extraction part consists of an image encoder and a text encoder. The input image is transformed into image embeddings through the image encoder while the input text prompt is encoded into text tokens through a text encoder.  
% As shown in the left part of Fig. \ref{framework}, the feature extraction component consists of an image encoder and a text encoder. 
The encoders are initialized with pre-trained weights and are partially frozen to reduce computation costs. % and prevent the issue of catastrophic forgetting.
In practice, we  utilize the frozen visual encoder ViT-G/14 from EVA-CLIP \cite{fang2023eva} and the text encoder from BLIP \cite{li2022blip}, which differs from the text and image encoders employed in CLIP \cite{radford2021learning}.

%\subsubsection{Multimodal Representation Learning}

% The network architecture of the proposed model. The InstructAIGCIQA  consists of three parts including (1) Feature Extraction part: the input text prompt first go through a LLM for prompt segmentation, then go to a text encoder to be transformed to text tokens while the input image go to image encoder to be transformed to image embeddings.
% % the image encoder is used to transform the image into image embeddings. 
% (2)Learning and Evaluating part: the text tokens and image embeddings then go through a Q-Former to interact with each other and learned by queries which then go though the multi-head quality regressor to be mapped to scores from three perspectives.
% (3)Instruction Tuning part: the learned queries then went to 
\subsubsection{Multimodal Representation Learning} Based on the extracted independent text and image features, a multi-modal Q-Former is applied to learn and extract multi-modal preference-aware representations, which employs a set of learnable query vectors to interact with image embeddings and text tokens using cross-attention and self-attention \cite{sun2024visual,duan2022saliency}. Specifically, these learnable queries interact with each other as well as the text-tokens through self-attention layers and interact with the image embeddings through cross-attention layers.
Different from BLIP-2 \cite{li2023blip2}, the Q-Former in our proposed model aims to extract the most informative quality-aware features from the partially frozen image and text encoders for the following regressors to output the desired scores consistent with human preferences from multiple perspectives.
\subsubsection{Score Regression}
After extracting quality-aware representations through the Q-Former, we map these features to the desired score spaces using the score regression module.
% We first apply the global average pooling (GAP) to the extracted quality representations. 
% This produces a feature vector with a dimension of K × 1, where K represents the number of final quality representations.
% Then two Fully Connected (FC) layers are used as the regressors to predict the human preference score. 
Three score regressors are used for quality, authenticity, and text-image correspondence score regression, respectively.  
%The structure and loss function of the three regressors are the same. 
However, the number of the regressors can be increased or decreased according to the number of the evaluation dimensions of the corresponding database, not limited to the three perspectives as mentioned above.
% we first apply an MLP to project the output quality-aware representations into the same dimension as the regression model input.
% Given one output learned query, 
To reduce the model complexity, the structure and loss function of the regressors remain the same.
% In each regressor, we first apply the global average pooling (GAP) to the extracted feature maps. 
% This produces a feature vector with a dimension of P × 1, where P represents the number of final feature maps. Then two Fully Connected (FC) layers are used as the regression model to predict the image quality. 
% Then, we use the L1 loss to guide the output score of each regressor. The loss function is as follows:
% %correlates well with its corresponding MOSs. 
% \begin{eqnarray}
% \begin{aligned}
% \label{loss_function}
% & Loss =\frac{1}{N}\sum_{i=1}^N \left| Pred(i) - Label(i)  \right|,
% \end{aligned}
% \end{eqnarray}
% where $Pred(i)$ is the score predicted by the regressor $i$ and $Label(i)$ is the corresponding ground-truth MOS derived from subjective experiments, $N$ is the total number of regressors used in the model.

\subsection{Understanding and Explaining with Instruction Tuning}

To gain a deeper understanding of the human visual preference for AIGIs and provide more detailed explanations for the multi-dimensional evaluation principle, we implement another Q-Former for vision-language instruction tuning. It inherits quality-aware information from the previous Q-Former, interacting with instructions and image embeddings to extract instruction-aware representations.
Then we apply a LLM to perform representation-guided visual question answering according to the given instruction.

\subsubsection{Instruction-aware Representation Extraction}
% The Q-Former for instruction tuning takes the same set of K learnable query embeddings as input to ensure the consistency with the first Q-Former. 
We incorporate the quality representation queries obtained from the previous Q-Former into the initial input queries of the instruction tuning Q-Former.
The two Q-Formers are connected through a zero convolution. The weights and biases of the zero convolution are initialized with zero but will be adjusted during the training process. 
By doing so, we can supplement the instruction tuning Q-Former with the quality-aware information from the first Q-Former, while avoiding adding initial noise to the pre-trained InstructBLIP \cite{instructblip} with the help of zero initialized convolution.
The instruction information is also given to the Q-Former after being encoded by a text encoder, enabling the Q-Former to extract instruction-aware visual representations from the output of the image encoder through cross-attention. 

% By doing so, we can supplement score information, which interact with the output of the image encoder through cross attention.
% The output of the Q-Former consists of K encoded instruction-aware representation vectors, one per query embedding, which then go through a linear projection and are subsequently fed to the frozen LLM.
% The two Q-Formers are connected with a unique type of convolutional layer called "zero convolution" as applied in ControlNet.

\subsubsection{Representation-guided Visual Question Answering}
We perform representation-guided degradation-related visual question answering by connecting the output of the instruction tuned Q-Former to a frozen LLM. The frozen LLM adopted in our model is Vicuna \cite{zheng2023judging}, a recently released decoder-only LLM finetuned from LLaMA \cite{touvron2023llama}. The connection part between the Q-Former and LLM is a fully-connected layer which adapts the instruction-aware visual representations to the input dimension of the LLM.
 Since the LLM is adapted to output different answers with instruction tuning, it can keep frozen to reduce computation while getting the ability of human visual preference explaining. 

\subsection{Training and Fine-tuning Strategy}

The proposed model undergoes a three-stage training process: (1) score regression pretraining stage, (2) vision-to-language instruction tuning stage, (3) feedback from the understanding module to evaluating module
During the first stage, the Q-Former is pre-trained for vision-language representation learning and image quality score regression. 
%with the partially frozen image encoder and the original text prompts as input.
% In the second stage, the output of the Q-Former is used as soft visual prompts for text generation with a frozen LLM.
In the second stage, we freeze the first Q-Former and initialize the instruction tuning Q-Former with a pre-trained InstructBLIP \cite{instructblip} model and further fine-tune it on the AIGCIQA2023+ database. 
For the third stage, the output of the explanation module is feedback to the score regression module to further improve the performance.
%with carefully crafted instructions to provide the most instructive information and use the output of the Q-Former as soft visual prompts for text generation with a frozen LLM.

% This stage aims to harvest the generative language capability of the LLM by connecting it to the Q-Former (with the frozen image encoder attached) and fine-tuning it with carefully crafted instructions.
% This stage aims to harvest the generative language capability of the frozen LLM by fine-tuning the Q-Former with carefully crafted instructions to provide the most instructive information.
 \begin{table*}[tbph]
% \vspace{-2mm}
\renewcommand\arraystretch{1.2}
\caption{Performance comparisons of the state-of-the-art IQA methods on the AIGCIQA2023+ database from three perspectives. The best performance results are marked in {\textcolor{red}{RED}} and the second-best performance results are marked in {\textcolor{mblue}{blue}}.}
  \label{t2}
  \centering
\footnotesize
\vspace{-4mm}
  \begin{tabular}{l||ccc|ccc|ccc|ccc}
  \multicolumn{1}{c}{} &  &  & \multicolumn{1}{c}{} &  &  & \multicolumn{1}{c}{} &  &  & \multicolumn{1}{c}{} &  &  & 
\tabularnewline
\hline 
\hline 
 & \multicolumn{3}{c|}{\textbf{Quality}} & \multicolumn{3}{c|}{\textbf{Authenticity}} & \multicolumn{3}{c|}{\textbf{Correspondence}} & \multicolumn{3}{c}{\textbf{Average}}\tabularnewline
\cline{2-13} \cline{3-13} \cline{4-13} \cline{5-13} \cline{6-13} \cline{7-13} \cline{8-13} \cline{9-13} \cline{10-13} \cline{11-13} \cline{12-13} \cline{13-13} 
 % & SRCC & PLCC & SRCC & PLCC & SRCC & PLCC & SRCC & PLCC & SRCC & PLCC & SRCC & PLCC \tabularnewline
 
 & \textbf{SRCC}  & \textbf{PLCC}  & \textbf{KRCC}     & \textbf{SRCC}  & \textbf{PLCC}  & \textbf{KRCC}  &  \textbf{SRCC}  & \textbf{PLCC}  & \textbf{KRCC}    & \textbf{SRCC}  & \textbf{PLCC}  & \textbf{KRCC}    \tabularnewline

\hline
    % \midrule
    % \textbf{Method}  & \textbf{SRCC}  & \textbf{PLCC}  & \textbf{KRCC}     & \textbf{SRCC}  & \textbf{PLCC}  & \textbf{KRCC}    & \textbf{SRCC}  & \textbf{PLCC}  & \textbf{KRCC}  & \textbf{SRCC}  & \textbf{PLCC}  & \textbf{KRCC}    \tabularnewline
    % \midrule

\textbf{NIQE} \cite{mittal2012making} & 0.5060  & 0.5218 & 0.3420 &0.3715  & 0.3954 &0.2453 & 0.3659  &0.3485 &0.2460 & 0.4145 & 0.4219 &  0.2778
    \tabularnewline
  \textbf{QAC} \cite{xue2013learning} &0.5328  &0.5991 &0.3644
&0.4009  &0.4428 &0.2673
&0.3526  &0.4062  &0.2414 
& 0.4288 & 0.4827 & 0.2910 \tabularnewline
    \textbf{BRISQUE} \cite{mittal2012no} &0.6239 &0.6389  &0.4291
&0.4705  &0.4796  &	0.3142
&0.4219  &0.4280  &	0.2865 
&0.5054 &0.5155 &  0.3433 \tabularnewline
%  \textbf{PRI-PSS} \cite{min2017blind} &0.3556  &0.4183 &0.2373
% &0.2409  &0.2625 &0.1583
% &0.2670  &0.2960  &0.1794 \tabularnewline
%     \textbf{PRI-LSSs} \cite{min2017blind} &0.5141  &0.5618 &0.3512
% &0.3721  &0.3998 &0.2460
% &0.3230  &	0.3473 &0.2160   \tabularnewline
%     \textbf{PRI-LSSn} \cite{min2017blind} &0.5245  &0.5935  &0.3523
% &0.3838  &0.5465  &0.2528
% &0.3655  &0.4594 &0.2474   \tabularnewline
   \textbf{BPRI}  \cite{min2017blind} &0.6301  &0.6889 &	0.4307
&0.4740 &0.5207   &	0.3144
&0.3946 &0.4346   &	0.2657
&0.4996 &0.5481   & 0.3369\tabularnewline
 
    \textbf{HOSA} \cite{xu2016blind} &0.6317 &	0.6561 &0.4311
&0.4716  &0.4985  &0.3101
&0.4101  &0.4252 &0.2765 
&0.5045  &0.5266 &0.3392\tabularnewline
\textbf{BMPRI} \cite{quality:BMPRI} &0.6732 &0.7492  &0.4661
&0.5273 &0.5756 &	0.3554 
&0.4419 &0.4827  &	0.3014  
&0.5475 &0.6025 & 0.3743\tabularnewline
%\cmidrule(r){1-13}

\textbf{Higrade-1} \cite{kundu2017large} &0.4849 &	0.4966 &0.3220
&0.4175& 	0.4181& 	0.2791
&0.3319& 	0.3379& 	0.2207
&0.4114&    0.4175&     0.2739\tabularnewline
\textbf{Higrade-2} \cite{kundu2017large}&0.2344& 0.3189 &	0.1568
&0.2654  &	0.3106 &	0.1742
&0.1756 &	0.2144 &0.1170
&0.2251 &   0.2813 & 0.1493\tabularnewline
% \cmidrule(r){1-10}
\hline
\textbf{CLIP Score} \cite{radford2021learning}& 0.2355  &	0.2629&	0.2629
&0.1456  &	0.0966&	0.1752
&0.2337  &	0.1578&   0.2729 
&0.2049  &  0.1724& 0.2370\tabularnewline
\textbf{BLIP Score} \cite{li2022blip}& 0.3485 &	0.2319&	0.3552 
&0.3191 &	0.2128  &	0.3275
&0.3784  &	0.2576 &   0.3882 
&0.3487 & 0.2341  &   0.3570\tabularnewline
\textbf{FLIP Score} \cite{li2023scaling}& 0.3424  &	0.2287&	0.3575
&0.2287 &	0.1717 &	0.2670 
&0.3561 &	0.2427  &   0.3822
&0.3091 &   0.2144  &   0.3356\tabularnewline
\textbf{Aesthetic Score} \cite{schuhmann2022laion}& 0.5879  &	0.4057&	0.5974
&0.5087  &	0.3473 &	0.5162
&0.4851 &	0.3299  &   0.4942
&0.5272 &   0.3610  &   0.5359\tabularnewline
\textbf{ImageReward Score} \cite{database/align:ImageReward}& 0.5153  &	0.5229&	0.3507
&0.4802  &	0.4836 &	0.3265
&0.5870   &   0.5911&	0.4094 
&0.5275   &   0.5325&   0.3622\tabularnewline

% \cmidrule(r){1-10}
\hline
\textbf{WaDIQaM-NR} \cite{bosse2017deep}& 0.4447 &	0.4996 &	0.3036
&0.3936  &	0.3906 &	0.2715
&0.3027  &	0.2810 	&0.2057
&0.3803  &  0.3904  & 0.2603
 \tabularnewline
 \textbf{CNNIQA} \cite{kang2014convolutional} &0.7160  &0.7937 &0.4955
&0.5958 &0.5734   &0.4085
&0.4758 &	0.4937 &0.3313
&0.5959 &0.6203  & 0.4118\tabularnewline
 \textbf{VGG16} \cite{simonyan2014very} &0.7961   &	0.7973 &0.5843
&0.6660    &0.6807  &0.4813
&0.6580    &0.6417  &0.4548  
&0.7067    &0.7066  &0.5068\tabularnewline
 \textbf{VGG19} \cite{simonyan2014very} &0.7733     &	0.8402 &0.5376 
&0.6674   &0.6565  &0.4843 
&0.5799     &0.5670  &0.4090     
&0.6735   & 0.6879   &0.4770\tabularnewline
 \textbf{Resnet18} \cite{he2016deep} &0.7583        &	0.7763  &0.5360
&0.6701     &0.6528     &0.4740 
&0.5979   &0.5564    &0.4165   
&0.6754   &0.6618    &0.4755\tabularnewline
 \textbf{Resnet34} \cite{he2016deep} &0.7229         &	0.7578   &0.4835    
&0.5998        &0.6285   &0.4325     
&0.7058     &0.7153    &0.5111 
&0.6762 &0.7005  &0.4757\tabularnewline
\textbf{MUSIQ} \cite{ke2021musiq} &  \bf\textcolor{mblue}{0.8421} & 0.8475 & \bf\textcolor{mblue}{0.6338}
& 0.7382 & 0.7278 & 0.5377 
& 0.7381 & 0.7216 & 0.5418
& 0.7728 & 0.7656 & 0.5711\tabularnewline
 \textbf{TreS} \cite{golestaneh2022no} & 0.8389 & \bf\textcolor{mblue}{0.8640} &0.6283 
  &\bf\textcolor{mblue}{0.7480}         &	\bf\textcolor{mblue}{0.7371}   &\bf\textcolor{mblue}{0.5511}    
&\bf\textcolor{mblue}{0.7401}        &\bf\textcolor{mblue}{0.7188}   &\bf\textcolor{mblue}{0.5448}        
&\bf\textcolor{mblue}{0.7757} & \bf\textcolor{mblue}{0.7733}  &\bf\textcolor{mblue}{0.5747}\tabularnewline
\textbf{CLIP-IQA+} \cite{wang2022exploring}&0.8300 & 0.8405 &0.6170 
& 0.7325 &0.7199 &0.5265
& 0.6484 &0.6424&0.4554
& 0.7370 &0.7343 &0.5330\tabularnewline

% \textbf{ImageReward(train)}& 0.8758&	0.8849 &	0.7490
% &0.8016  &	0.7990 &	0.6017
% &0.8141  &	0.8004 &   0.6119  \tabularnewline
\hdashline
% \textbf{Proposed-single}& \bf\textcolor{mblue}{0.8769} &	\bf\textcolor{mblue}{0.8874} &	\bf\textcolor{mblue}{0.6802}
% &\bf\textcolor{mblue}{0.8209} &	\bf\textcolor{mblue}{0.8159} &	\bf\textcolor{mblue}{0.6236} 
% &\bf\textcolor{mblue}{0.8342} &   \bf\textcolor{mblue}{0.8191} &	\bf\textcolor{red}{0.6357}  \tabularnewline
\rowcolor{gray!20}\textbf{MINT-IQA (Ours)}& \bf\textcolor{red}{0.8801} &	\bf\textcolor{red}{0.8870} &	\bf\textcolor{red}{0.6841}
&\bf\textcolor{red}{0.8229} &	\bf\textcolor{red}{0.8127} &	\bf\textcolor{red}{0.6223}
&\bf\textcolor{red}{0.8226} &   \bf\textcolor{red}{0.8055} &	\bf\textcolor{red}{0.6231}  
&\bf\textcolor{red}{0.8419} &\bf\textcolor{red}{0.8351} & \bf\textcolor{red}{0.6432}\tabularnewline
\rowcolor{gray!20}\textit{Improvement} 
& + 3.8\%
& + 2.3\% & + 5.0\% &+ 7.5\%  
& + 7.6\%
& + 7.1\% & + 8.2\% &+ 8.7\%  
& + 7.8\% 
& + 6.6\% & + 6.2\% &+ 6.9\% 
 \tabularnewline
\bottomrule
  \end{tabular}\label{tab:AIGCIQA2023-results}
 \vspace{-2mm}
\end{table*}
 
\subsubsection{Score Regression Pre-training}
%  Also, we propose a balanced sampling strategy to synchronize learning progress across datasets.
%  The models are trained with the standard language modeling loss to directly generate the response given the instruction.
There are three types of image annotations in current AIGC IQA databases including MOS, pairs, and rankings.
For the MOS input format, we simply use the $L1$ loss to optimize the training process. The loss function is formulated as follows:
%correlates well with its corresponding MOSs. 
\vspace{-2mm}
\begin{eqnarray}
\begin{aligned}
\label{loss_function}
& \mathcal{L} =\frac{1}{N}\sum_{i=1}^N \space \left| \space Q_{\text{predict}}(i) - Q_{\text{label}}(i)  \right|,
\end{aligned}
\vspace{-3mm}
\end{eqnarray}
where $Q_{\text{predict}}(i)$ is the score predicted by the regressor $i$ and $Q_{\text{label}}(i)$ is the corresponding ground-truth MOS collected from subjective experiments, $N$ is the total number of regressors used in the model.
For the annotation format of pairs, we use the cross-entropy loss to optimize the network. Suppose $\{P_b,P_w\}$ is a pair of images for comparison and $P_b$ is the better one while $P_w$ is the worse, the loss function can be formulated as:
\begin{equation} \label{eq:loss}
\begin{split}
    \mathcal{L}(\theta) = - \mathbb{E}_{(T, P_b, P_w) \sim \mathcal{D}}[\mathop{\log}(\sigma(f_\theta(T, P_b) - f_\theta(T, P_w)))],
\end{split}
\end{equation}
where $f_\theta(T,P)$ is predicted preference score for prompt $T$ and generated image $P$.
For the ranking annotations, the images from the best to the worst can be denoted as $P_1, P_2, ..., P_k$. We can arrange and combine them into the following pairs: $\{P_1,P_2\},  \{P_1,P_3\}, ..., \{P_{k-1},P_k\}$.
If there are $k$ ranking images with the same prompt $T$, we can obtain a maximum of $C_k^2$ comparison pairs.
Then the cross-entropy loss function used for pair format can be applied for the ranking format.%if there are no ties between two images. Then, we use the same cross-entropy loss as the form of pairs input.
% Training AICIQA-LLM is of no ease. 
% We observe rapid convergence and consequent overfitting, which negatively impacts its performance. To address this, we freeze the parameters of certain backbone transFormer layers. 
% We have found that fixing an appropriate number of layers improves its performance.
% AICIQA-LLM also exhibits sensitivity to training hyperparameters, such as learning rate and batch size. We perform a meticulous grid search based on the validation set to determine the optimal values.
\begin{figure}[t]
% \vspace{-3mm}
	\centering
	\includegraphics[width=1\linewidth]{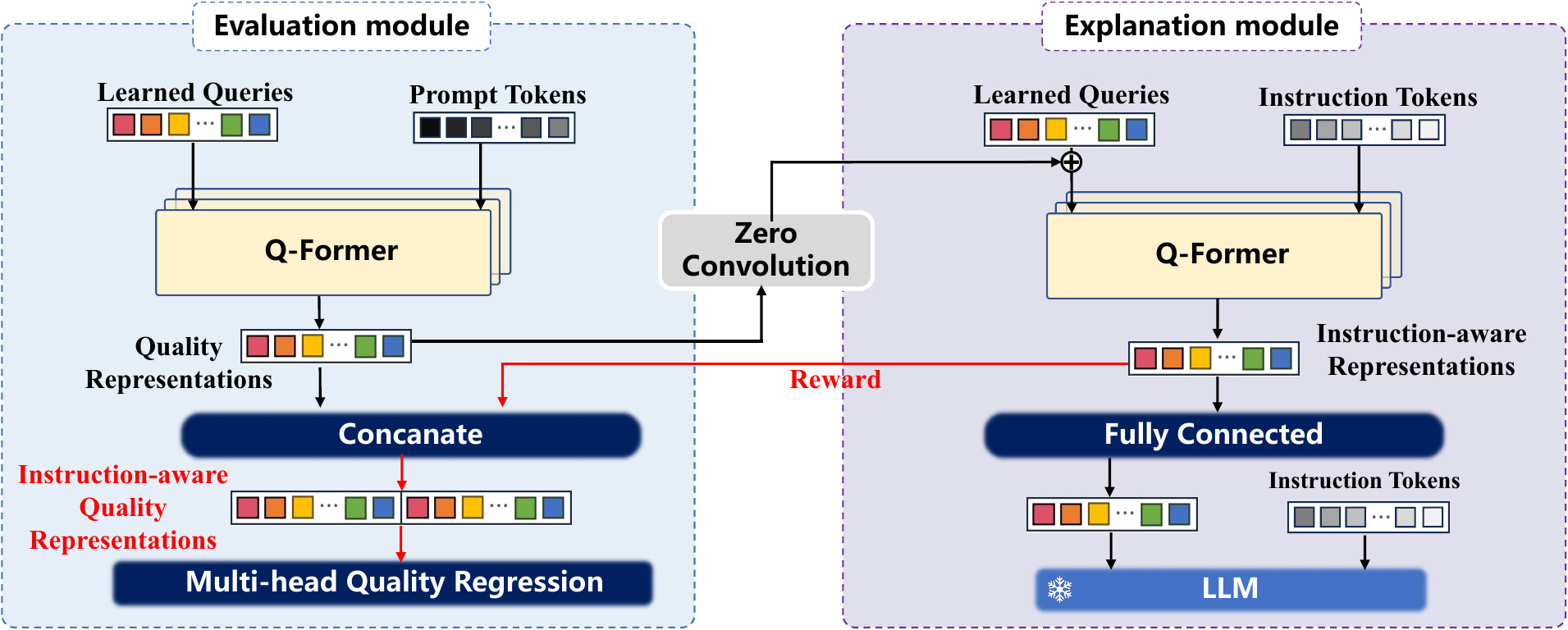}
 \vspace{-7mm}
	\caption{Feedback from the explanation module to evaluation module. (Improvement in \textcolor{red}{RED}.)}
	\label{ablation2}
 \vspace{-3mm}
\end{figure}
\subsubsection{Vision-to-Language Instruction Tuning}
% Similar to RM training for language model of previous works, we formulate the preference annotations as rankings. 
% We have $k\in[4,9]$ images ranked for the same prompt $T$ (the best to the worst are denoted as $x_1, x_2, ..., x_k$) and get at most $C_k^2$ comparison pairs if no ties between two images. 
 % fine-tune the MINT-IQA based on the MINT-IQA database,
To further enhance the explanation capabilities of our model, we propose a vision-language instruction tuning strategy.
% We first froze the previous Q-Former to reduce computational complexity and avoid affecting the previous score prediction procedure, we froze the previous Q-Former.
We initialize the weights of the instruction tuning Q-Former with pretrained InstructBLIP \cite{instructblip}, which empowers the model with initial ability of complex visual scene understanding. To further enhance the degradation-awareness capabilities of our model, we fine-tune it on our AIGCIQA2023+ database to adapt the frozen LLM to give fine-grained explanations for the quality assessment basis of each AIGI.
The model is trained using the standard language modeling loss. We freeze the first Q-Former during training to reduce computational complexity and avoid affecting the previous score prediction procedure.
% with carefully crafted instructions.
% to provide the most instruction-aware visual features to the frozen LLM.
% To reduce computational complexity and avoid affecting the previous score prediction procedure, we froze the previous Q-Former during training.
% The model is trained using the standard language modeling loss.
This instruction tuning strategy enables the MINT-IQA model to further attain the powerful capability for human visual preference explanation while maintaining the original complex visual scene understanding ability. 

		\begin{table}[!tp]
  % \vspace{-2mm}
  \renewcommand\arraystretch{1.1}
			\caption{Performance comparisons on the AGIQA-3K database. The best results are marked in {\textcolor{red}{RED}} and the second results are marked in {\textcolor{mblue}{blue}}.}
   \footnotesize
			\label{tab:perception}
			\centering
   \vspace{-2mm}
				\begin{tabular}{llccc}
					\toprule
					\multicolumn{1}{l}{\textbf{Dimension} }  &\textbf{Method}                      & \textbf{SRCC} & \textbf{PLCC}  & \textbf{KRCC}      \\   \midrule
                            \multirow{17}{*}{Quality}&CEIQ \cite{quality:CEIQ}                    & 0.3228  & 0.4166 & 0.2220 \\
					 &DSIQA \cite{quality:DSIQA}                   & 0.4955  & 0.5488 & 0.3403  \\
					 &NIQE \cite{mittal2012making}                    & 0.5623  & 0.5171 & 0.3876 \\
				&Sisblim \cite{quality:sisblim}                 & 0.5479  & 0.6477 & 0.3788  \\ 
					 &FID \cite{heusel2017gans}                     & 0.1733 & 0.1860  & 0.1158  \\
					&ICS \cite{quality:ics}                     & 0.0931  & 0.0964 & 0.0626  \\
					&KID \cite{quality:kid}                     & 0.1023  & 0.0786 & 0.0692  \\ 
					 &BMPRI \cite{quality:BMPRI}                   & 0.6794 & 0.7912  & 0.4976\\
					&GMLF \cite{quality:GMLF}                    & 0.6987  & 0.8181 & 0.5119  \\
					&Higrade \cite{kundu2017large}                 & 0.6171  & 0.7056 & 0.4410  \\ 
					&DBCNN \cite{quality:DBCNN}                   & 0.8207  & 0.8759 & 0.6336\\
					% CLIPIQA \cite{radford2021learning}                & {\bf\textcolor{mblue}{0.8426}}  & 0.8053 & 0.6468 \\
					&CNNIQA \cite{kang2014convolutional}                  & 0.7478  & 0.8469 & 0.5580\\
					&HyperNet \cite{quality:HyperNet}                & 0.8355 & 0.8903  & 0.6488\\ 
                    &CLIP-IQA+~\cite{wang2022exploring} & 0.8428 & 0.8879 & 0.6556\\
                    % Q-Bench (InternLM) &0.732&0.775& -\\
                    &CLIP-AGIQA~\cite{fu2024vision} & \bf\textcolor{mblue}{0.8747} & \bf\textcolor{mblue}{0.9190} & \bf\textcolor{mblue}{0.6976} \\
                    
       \cdashline{2-5}
     \rowcolor{gray!20}&\textbf{MINT-IQA (Ours)}      & {\bf\textcolor{red}{0.8919}} & {\bf\textcolor{red}{0.9328}}  & {\bf\textcolor{red}{0.7224}}\\
     \rowcolor{gray!20}&\textit{Improvement} 
& + 1.7\%
& + 1.4\% & + 2.5\% 
 \\
     \hline
     \multirow{7}{*}{Alignment}&
     CLIP \cite{radford2021learning}                                          & 0.5972 & 0.6839  & 0.4591 \\
					
					&HPS \cite{database/align:HPS}                                          & 0.6349  & 0.7000 & 0.4580  \\
					&PickScore \cite{database/align:PickAPic}                                    & 0.6977 & 0.7633  & 0.5069\\
     &ImageReward \cite{database/align:ImageReward}                                  & 0.7298& 0.7862 & 0.5390\\
                        &StairReward  \cite{database/agiqa}                                   & {\bf\textcolor{mblue}{0.7472}}  & {\bf\textcolor{mblue}{0.8529}}& {\bf\textcolor{mblue}{0.5554}}
                            \\  
                             \cdashline{2-5}
                         % Proposed-single                                  & {\bf\textcolor{mblue}{0.8749}} & {\bf\textcolor{mblue}{0.9282}} & {\bf\textcolor{mblue}{0.7036}} \\
					\rowcolor{gray!20} &\textbf{MINT-IQA (Ours)}                                  & {\bf\textcolor{red}{0.8749}} & {\bf\textcolor{red}{0.9282}}  & {\bf\textcolor{red}{0.7036}}\\ 
                     \rowcolor{gray!20}&\textit{Improvement} 
& + 12.8\%
& + 7.5\% & + 14.8\% 
 \\
					\hline
    
			\end{tabular}
      \vspace{-2mm}
			\end{table}

\subsubsection{Feedback from the explanation module to evaluating module}

 To further strengthen the connection between the two frameworks and help predict more accurate scores, we take the instruction-aware representations from the second framework as a reward. As shown in Fig. \ref{ablation2}, we concanate the instruction-aware representations and the quality-aware representations as the input for multi-head quality regression, and then fine-tune the three regressors respectively with corresponding instructions. For example, when finetuning the quality regressor, the instruction is “how is the quality of the image”. The instruction-aware representations according to this instruction are highly quality-aware. Consequently, taking these representations from the second framework as a reward to the first framework leads to more accurate quality score prediction. As shown in Table \ref{tab:9d}, the model gains better performance through the feedback.

\section{Experiment}

In this section, we conduct extensive experiments to evaluate the performance of our proposed model.
We first present the experimental protocol in detail. Then we launch experiments to assess the IQA performance of our model compared to current state-of-the-art IQA models in predicting human preference based on three AIGC IQA databases and three traditional IQA databases.
To quantitatively measure the accuracy of our model in explaining human preference for AIGIs, we propose  Visual Question Answering Accuracy ($\text{VQA}_\text{acc}$) for the evaluation of the human preference explanation task. We launch further cross-dataset experiments to test the generalizability of proposed model.
% from the perspectives of quality, authenticity, and correspondence 
% based on three AIGC databases.
% To evaluate the generalizability of our proposed model, we further launch cross-dataset evaluation experiments.
Finally, we conduct ablation experiments to evaluate the efficiency of our proposed components.
% Finally, we report the comparison results between the proposed model and other state-of-the-art IQA models on both AIGC databases and traditional IQA databases.
% In this section, we aim to evaluate the performance of our proposed model compared to other state-of-the-art IQA models.
% We first introduce the test databases and existing IQA models, followed by the training and evaluation protocols.
% Next, we launch baseline experiment to evaluate the performance of current state-of-the-art IQA models in predicting human preference from the perspectives of quality, authenticity, and correspondence on the AIGCIQA2023 dataset.
% We then evaluate the generality and pervasiveness of our proposed model by conducting several comparison experiments on two other related AIGC databases and three traditional IQA databases.
% % We then introduce the other two related AIGC databases and three traditional IQA databases and then present the experimental protocol in detail. 
% We report the comparison results between the proposed model and other state-of-the-art IQA models on both AIGC databases and traditional IQA databases.
% \subsection{Basic Experiment on AIGCIQA2023}
% Based on the proposed theory \textit{i.e.} assessing AIGIs from perspectives of quality, authenticity and correspondence, we establish a new benchmark to evaluate the performance of current state-of-the-art IQA models and show that the current mainstream evaluation metrics are not well correlated with human preferences.

% \input{tables/3k}

\subsection{Experimental Protocol}
\begin{table}[!tp]
% \vspace{-2mm}
\renewcommand\arraystretch{1.1}
  \caption{Human preference prediction metric performance comparisons on the ImageReward database. Preference accuracy is conducted based on the test set with 466 prompts (6,399 comparisons). The best results are marked in {\textcolor{red}{RED}} and the second results are marked in {\textcolor{mblue}{blue}}. }\label{imagereward}
  \centering\footnotesize
  \vspace{-2mm}
  \begin{tabular}{lr}
    \toprule
      \textbf{Method} & \bf{Accuracy} \\
    \midrule
      CLIP Score \cite{radford2021learning} & 54.82  \\
      Aesthetic Score \cite{schuhmann2022laion} & 57.35  \\
      BLIP Score \cite{li2022blip} & 57.76 \\
      ImageReward \cite{database/align:ImageReward}& \bf\textcolor{mblue}{65.14}  \\
    \hdashline
\rowcolor{gray!20}      \textbf{MINT-IQA (Ours)}  & \bf\textcolor{red}{66.46} \\
\rowcolor{gray!20}\textit{Improvement} 
& + 1.3\%\\

    \hline
  \end{tabular}
  \vspace{-1mm}
\end{table}

% \subsubsection{Test Databases}
We assess the performance of our proposed method on six IQA databases including three AIGC IQA databases (AIGCIQA2023 \cite{wang2023aigciqa2023}, AGIQA-3K \cite{database/agiqa}, ImageReward \cite{database/align:ImageReward})
and three traditional IQA databases (KonIQ-10k \cite{hosu2020koniq}, SPAQ \cite{fang2020perceptual}, AVA \cite{ava}).
 % \subsubsection{Evaluation Criteria}
To evaluate the correlation between the predicted scores and the corresponding ground-truth MOSs, we utilize the following three correlation coefficients as performance evaluation criteria, including Spearman Rank Correlation Coefficient (SRCC), Pearson Linear Correlation Coefficient (PLCC), and Kendall’s Rank Correlation Coefficient (KRCC).

% \subsubsection{Experimental Setup}
% All the IQA models are validated on the proposed AIGCIQA2023 database. 
%For SVR-based models, the repeating time is 1,000, implemented by LIBSVM  with radial basis function (RBF) kernel. 
% The repeating time is 10, the training epochs are 50 with an initial learning rate of 0.0001 and batch size of 4 on AIGCIQA2023 database.
Some baseline results are evaluated by ourselves. For traditional handcrafted-based models, they are directly evaluated on the corresponding databases.
For vision-language pre-training models, we simply load the pre-trained weights for inference. 
CLIP score and BLIP score are calculated directly as cosine similarity between text and image embeddings.
% CLIP \cite{radford2021learning}, BLIP \cite{li2022blip}, FLIP \cite{li2023scaling}, Aesthetic score \cite{schuhmann2022laion} and ImageReward \cite{database/align:ImageReward} are based on a pre-trained ViT-L/14 CLIP \cite{radford2021learning} image encoder. We simply load the pre-trained weights for inference. 
For deep learning-based models, we use the same training and testing split as the previous literature.
For our proposed model, the number of regressors in our model is flexible, corresponding to the varying number of evaluation dimensions in the test databases. 
Since traditional IQA datasbases have only one dimension for assessment, only one regressor is used for score prediction. 
The models are implemented with PyTorch and trained on a 40GB NVIDIA RTX A6000 GPU with batch size of 8. The initial learning rate is set to 1e-5, and decreased using the cosine annealing strategy. We employ Adam optimizer with $\beta_1 =0.9$ and $\beta_2 = 0.999$.
 
% All the results from our method are averaged across three runs. 

% \subsubsection{Performance Discussion}

% The performance results of the state-of-the-art IQA models mentioned above on AIGCIQA2023 database are exhibited in Table \ref{tab:AIGCIQA2023-results}, from which we can make several conclusions:
% \begin{itemize}
% \item The handcrafted-based methods achieve poor performance on the whole database, which indicates the extracted handcrafted features are not effective for modeling the quality representation of AIGIs. This is because most employed handcrafted features of these methods are based on the prior knowledge learned from NSIs, which are not effective for evaluating AIGIs.

% \item The deep learning-based methods achieve relatively more competitive performance results on three evaluation perspectives. However, they are still far away from satisfactory.

% \item Most of the IQA models achieve better performance on quality evaluation and worse on text-image correspondence score assessment.
% The reason is that the text prompts for image generation are not utilized for the IQA model training. 
% This makes it more challenging for the IQA models to extract relation features from AIGIs, which inevitably leads to performance drops.
% \end {itemize}
% The ablation studies are conducted to validate the effectiveness
% of each module in the proposed model.
% \vspace{-3mm}
 \subsection{Performance on AIGC IQA Databases}

 We evaluate the performance of our proposed metric, MINT-IQA, on three recently released AIGC IQA databases including AIGCIQA2023 \cite{wang2023aigciqa2023}, AGIQA-3K \cite{database/agiqa}, and ImageReward \cite{database/align:ImageReward}. 
% Then, we select a wide range of benchmarks for comparison. 
The performance comparison results are demonstrated in Tables \ref{tab:AIGCIQA2023-results}-\ref{imagereward}.
% \textbf{Results on AIGCIQA2023:}\space
Handcrafted-based models show poor performance on AIGC IQA databases, indicating that the handcrafted features mainly targeted for natural images are ineffective for evaluating AIGIs.
% in accurately representing the quality of AIGIs.
% This ineffectiveness can be attributed to the fact that the majority of handcrafted features are derived from poor knowledge acquired from natural images, which may not be suitable for evaluating AIGIs. 
 % \begin{table}[h!]
% \centering
% \caption{Summary of IQA datasets.}
% \resizebox{2.1 in}{!} {
% \begin{tabular}{cccc}
% \hline 
% \multirow{2}{*}{Databases} & \# of Dist. & \# of Dist.  & Distortions \tabularnewline
%  & Images & Types & Type\tabularnewline
% \hline 
% LIVE & 799 & 5 & synthetic\tabularnewline
% CSIQ & 866 & 6 & synthetic\tabularnewline
% TID2013 & 3,000 & 24 & synthetic\tabularnewline
% KADID & 10,125 & 25 & synthetic\tabularnewline
% CLIVE & 1,162 & - & authentic\tabularnewline
% KonIQ & 10,073 & - & authentic\tabularnewline
% LIVEFB & 39,810 & - & authentic\tabularnewline
% \hline 
% \end{tabular}
% }
% \label{TB0}

% \end{table}

\begin{table*}[t!]
\centering
\renewcommand\arraystretch{1.15}
\caption{Comparison of \textit{MINT-IQA} v.s. state-of-the-art NR-IQA algorithms on traditional synthetically and authentically distorted datasets.
The best performance results are marked in \textcolor{red}{RED} and the second-best performance results are marked in \textcolor{mblue}{blue}}.
\vspace{-4mm}
\resizebox{6.8 in}{!} {

\begin{tabular}{l||cc|cc|cc|cc|cc|cc}
\multicolumn{1}{c}{} &  & \multicolumn{1}{c}{} &  & \multicolumn{1}{c}{} &  & \multicolumn{1}{c}{} &  & \multicolumn{1}{c}{} &  & \multicolumn{1}{c}{} &  & 
\tabularnewline
\hline 
\hline 
 & \multicolumn{2}{c|}{LIVE \cite{sheikh2006statistical}} & \multicolumn{2}{c|}{CSIQ \cite{larson2010most}} & \multicolumn{2}{c|}{TID2013 \cite{ponomarenko2015image}} & \multicolumn{2}{c|}{KADID \cite{lin2019kadid}} & \multicolumn{2}{c|}{CLIVE \cite{ghadiyaram2015massive}} & \multicolumn{2}{c}{KonIQ-10k \cite{hosu2020koniq}} \tabularnewline
\cline{2-13} \cline{3-13} \cline{4-13} \cline{5-13} \cline{6-13} \cline{7-13} \cline{8-13} \cline{9-13} \cline{10-13} \cline{11-13} \cline{12-13} \cline{13-13} 
 & SRCC & PLCC & SRCC & PLCC & SRCC & PLCC & SRCC & PLCC & SRCC & PLCC & SRCC & PLCC \tabularnewline
\hline 
% HFD\cite{wu2017hierarchical} & \textbf{0.971} & 0.951 & 0.890 & 0.842 & 0.681 & 0.764 & - & - & - & - & - & - \tabularnewline
PQR\cite{zeng2017probabilistic}  & 0.965 & \bf\textcolor{mblue}{0.971}  & 0.873 & 0.901  & 0.849 & 0.864 & - & -  & 0.808 & 0.836 & 0.880 & 0.884 \tabularnewline
DIIVINE\cite{saad2012blind} & 0.892  & 0.908 & 0.804 & 0.776 & 0.643 & 0.567 & 0.413  & 0.435 & 0.588 & 0.591 & 0.546 & 0.558 \tabularnewline
BRISQUE\cite{mittal2012no} & 0.929 & 0.944 & 0.812 & 0.748 & 0.626 & 0.571 & 0.528 & 0.567 & 0.629  & 0.629  & 0.681 & 0.685\tabularnewline
ILNIQE\cite{zhang2015feature} & 0.902 & 0.906 & 0.822 & 0.865 & 0.521 & 0.648 & 0.534  & 0.558 & 0.508 & 0.508 & 0.523 & 0.537 \tabularnewline
BIECON\cite{kim2017fully} & 0.958 & 0.961 & 0.815 & 0.823 & 0.717 & 0.762 & 0.623 & 0.648 & 0.613 & 0.613 & 0.651 & 0.654 \tabularnewline
MEON\cite{ma2017end} & 0.951 & 0.955 & 0.852 & 0.864 & 0.808 & 0.824 & 0.604 & 0.691 & 0.697 & 0.710 & 0.611 & 0.628 \tabularnewline
WaDIQaM\cite{bosse2017deep} & 0.960 & 0.955 & 0.852 & 0.844 & 0.835 & 0.855 & 0.739 & 0.752 & 0.682 & 0.671 & 0.804 & 0.807 \tabularnewline
DBCNN\cite{quality:DBCNN} & 0.968 & \bf\textcolor{mblue}{0.971} & 0.946 & 0.959 & 0.816 & 0.865 & 0.851 & 0.856 & 0.869 & 0.869 & 0.875 & 0.884 \tabularnewline
TIQA\cite{you2020transformer} & 0.949 & 0.965 & 0.825 & 0.838 & 0.846 & 0.858 & 0.850 & 0.855 & 0.845 & 0.861 & 0.892 & 0.903 \tabularnewline
MetaIQA\cite{zhu2020metaiqa} & 0.960 & 0.959 & 0.899 & 0.908 & 0.856 & 0.868 & 0.762 & 0.775 & 0.835 & 0.802 & 0.887 & 0.856 \tabularnewline
P2P-BM\cite{ying2019patches} & 0.959 & 0.958 & 0.899 & 0.902 & 0.862 & 0.856 & 0.840 & 0.849 & 0.844 & 0.842 & 0.872 & 0.885 \tabularnewline
HyperIQA \cite{su2020blindly} & 0.962 & 0.966 & 0.923 & 0.942 & 0.840 & 0.858 & 0.852 & 0.845 & 0.859 & 0.882 & 0.906 & 0.917 \tabularnewline
CONTRIQUE \cite{madhusudana2022image}& 0.960 & 0.961 & 0.942 & 0.955 & 0.843 & 0.857 & 0.934 & 0.937 & 0.845 & 0.857 & 0.894 & 0.906 \tabularnewline
% MUSIQ \cite{ke2021musiq} & 0.837 & 0.818 & 0.697 & 0.766 & - & - & 0.572 & 0.584 & 0.785 & 0.828 & 0.915 & 937
TReS \cite{golestaneh2022no}& 0.969 & 0.968 & 0.922 & 0.942 & \bf\textcolor{mblue}{0.863} & \bf\textcolor{mblue}{0.883} & 0.859 & 0.858 & 0.846 & 0.877 & 0.915 & \bf\textcolor{mblue}{0.928} \tabularnewline
Re-IQA \cite{saha2023re}& \bf\textcolor{mblue}{0.970} & \bf\textcolor{mblue}{0.971} & \bf\textcolor{mblue}{0.947} & \bf\textcolor{mblue}{0.960} & 0.844 & 0.880 & 0.885 & 0.892 & 0.840 & 0.854 & 0.914 & 0.923 \tabularnewline
LIQE \cite{zhang2023blind} & \bf\textcolor{mblue}{0.970} & 0.951 & 0.936 & 0.939 & - & - & \bf\textcolor{mblue}{0.930} & \bf\textcolor{mblue}{0.931} & \bf\textcolor{mblue}{0.904} & \bf\textcolor{mblue}{0.910} & \bf\textcolor{mblue}{0.919} & 0.908 \tabularnewline
\hdashline
\rowcolor{gray!20}MINT-IQA (Ours) &  \bf\textcolor{red}{0.971} & \bf\textcolor{red}{0.973} & \bf\textcolor{red}{0.976} & \bf\textcolor{red}{0.981} & \bf\textcolor{red}{0.894} & \bf\textcolor{red}{0.899} & \bf\textcolor{red}{0.964} & \bf\textcolor{red}{0.966} & \bf\textcolor{red}{0.908} & \bf\textcolor{red}{0.925} & \bf\textcolor{red}{0.927} & \bf\textcolor{red}{0.945} \tabularnewline
\rowcolor{gray!20}\textit{Improvement} 
& + 0.1\%
& + 0.2\% & + 2.9 \% &+ 2.1\%  
& + 3.1\%
& + 1.6\% & + 3.4\% &+ 3.5\%  
& + 0.4\% 
& + 1.5\% & + 0.8\% &+ 1.7\% 
 \tabularnewline

\hline 
\multicolumn{1}{c}{} &  & \multicolumn{1}{c}{} &  & \multicolumn{1}{c}{} &  & \multicolumn{1}{c}{} &  & \multicolumn{1}{c}{} &  & \multicolumn{1}{c}{} &  &  \tabularnewline
\end{tabular}

}
\vspace{-3mm}
\label{TB1}
\end{table*}

\begin{table}[!tp]
\vspace{-2mm}
\renewcommand\arraystretch{1.1}
\caption{Results on SPAQ dataset. The best performance results are marked in \textcolor{red}{RED} and the second-best performance results are marked in \textcolor{mblue}{blue}.}
\vspace{-4mm}
\begin{center}
\footnotesize
\begin{tabular}{lccc}\toprule
\textbf{Method} &\textbf{SRCC} &\textbf{PLCC} \\\midrule
DIIVINE \cite{moorthy2011blind} &0.599 &0.600 \\
BRISQUE \cite{mittal2012no} &0.809 &0.817 \\
CORNIA \cite{ye2012unsupervised} &0.709 &0.725 \\
QAC \cite{xue2013learning} &0.092 &0.497 \\
ILNIQE \cite{zhang2015feature} &0.713 &0.721 \\
FRIQUEE \cite{ghadiyaram2017perceptual} &0.819 &0.830 \\
DBCNN \cite{quality:DBCNN} &0.911 &0.915 \\
Fang \cite{fang2020perceptual} (w/o extra info) &0.908 &0.909 \\
MUSIQ \cite{ke2021musiq} &0.917 &0.921 \\
Tres \cite{golestaneh2022no}& 0.915 & 0.918 \\
LIQE \cite{zhang2023blind}&0.881 & -\\
Re-IQA \cite{saha2023re}&\bf\textcolor{mblue}{0.918} &\bf\textcolor{mblue}{0.925}\\
Q-Bench (InternLM) \cite{wu2023q} &0.730&0.750
\\\hdashline
\rowcolor{gray!20} \textbf{MINT-IQA (Ours)}  &\bf\textcolor{red}{0.927} &\bf\textcolor{red}{0.932} \\
 \rowcolor{gray!20}\textit{Improvement} 
& + 0.9\%
& + 0.7\% \\

\hline
\end{tabular}
\end{center}
\vspace{-4mm}
\end{table}
On the other hand, deep learning-based methods achieve relatively better results.
However, they are still far away from satisfactory.
MINT-IQA attains state-of-the-art performance across all three AIGC IQA datasetbases, substantially outperforming current IQA models. As shown in Table \ref{tab:AIGCIQA2023-results}, our proposed MINT-IQA model achieves the best performance  on our AIGCIQA2023+ from all three perspectives, which demonstrates the effectiveness of 
MINT-IQA in evaluating human preferences from multiple perspectives.
% It achieves best performance in predicting scores from all of the three perspectives on AIGCIQA2023 \cite{database/aigciqa2023}. 
Additionally, as demonstrated in Table \ref{tab:perception}, our model outperforms other methods in measuring the quality perception and text-image alignment on AGIQA-3K \cite{database/agiqa}.
This highlights the versatility of our model in understanding human preference for AIGIs in terms of both quality perception perspective and alignment perspective.
% Besides, joint training is more effective than single training, for joint training can make the model learn the feature representation from more diverse image content and distortions in other perspectives.
% MINT-IQA consistently surpasses its original backbone, BLIP-2, by a significant margin across all LLMs,demonstrating the effectiveness of vision-language instruction tuning.
% \textbf{Results on AGIQA-3K:}\space
%  Our method outperforms previous methods by a large margin in measuring both the perception and alignment dimensions in AGIQA-3K \cite{database/agiqa} . This highlights the versatility of our model in capturing not only the quality perception level but also the semantic level for AIGIs.
% \textbf{Results on ImageReward:}\space
Our proposed method also shows superior performance on the ImageReward \cite{database/align:ImageReward} dataset compared to the ImageReward model, as shown in Table \ref{imagereward}, even though we change the training data format from a single image and its MOS to a pair of images and their comparisons.
This further manifests the generalization ability of our model.
 %, even it is used in other models. 
 % MINT-IQA is compatible with any type of transformer models that accept input as a sequence of tokens.

% It is worth mentioning that due to the reasonable disassembly of the prompt, our StairReward far
% outperforms other methods in predicting the alignment of long prompts, thus taking the lead in the alignment index of the
% entire AGIQA-3K. Our approach is as universal as ever on large-scale datasets.
% \input{tables/AIGC}

 % \vspace{-2mm}
\subsection{Performance on Traditional IQA Databases}

\begin{table}[!tp]
\footnotesize
\vspace{-2mm}
\renewcommand\arraystretch{1.1}
\caption{Results on AVA dataset. The best performance results are marked in \textcolor{red}{RED} and the second-best performance results are marked in \textcolor{mblue}{blue}.}
\label{tab:ava-results}
\vspace{-4mm}
\begin{center}
\setlength\tabcolsep{2.5pt}
\begin{tabular}{lcccc}
\toprule
\textbf{Method} & \textbf{SRCC}& \textbf{PLCC} \\\midrule
Kong  \cite{kong2016photo} & 0.558 & - \\
% AMP \cite{murray2017deep}  & 0.709 & - \\
NIMA (VGG16) \cite{talebi2018nima}  & 0.592 & 0.610 \\
NIMA (Inception-v2) \cite{talebi2018nima}  & 0.612 & 0.636 \\
Zeng  (ResNet101) \cite{zeng2019unified}  & 0.719 & 0.720 \\
Hosu \cite{hosu2019effective} (\textbf{20} crops) & \bf\textcolor{mblue}{0.756} & \bf\textcolor{mblue}{0.757} \\
AFDC + SPP (single warp) \cite{chen2020adaptive}  & 0.648 & - \\
AFDC + SPP (\textbf{4} warps) \cite{chen2020adaptive}  & 0.649 & 0.671 \\
MUSIQ \cite{ke2021musiq} & 0.726 & 0.738 \\\hdashline
 \rowcolor{gray!20}\textbf{MINT-IQA (Ours)}  & \bf\textcolor{red}{0.776} & \bf\textcolor{red}{0.783} \\
 \rowcolor{gray!20}\textit{Improvement} 
& + 2.0\%
& + 2.6\% \\
\hline
\end{tabular}
\vspace{-4mm}
\end{center}

\end{table}
% \vspace{-1mm}
To further emphasize the effectiveness and generality of our proposed model, we also test it on eight traditional large-scale image quality datasets including LIVE \cite{sheikh2006statistical}, CSIQ \cite{larson2010most}, TID2013 \cite{ponomarenko2015image}, and CLIVE \cite{ghadiyaram2015massive},
 KADID \cite{lin2019kadid}, KonIQ-10k \cite{kong2016photo}, SPAQ \cite{fang2020perceptual} and AVA \cite{ava}.
% two technical quality datasets ( KonIQ-10k \cite{hosu2020koniq} and SPAQ \cite{fang2020perceptual} ) and one aesthetic quality dataset (AVA \cite{ava}).
 From Tables \ref{TB1}-\ref{tab:ava-results}, we first observe that the proposed model achieves the best performance on all eight traditional IQA databases. Our method outperforms the current state-of-the-art methods in terms of both SRCC and PLCC. The results indicate that the proposed model has more powerful representation learning abilities in assessing the quality of traditional natural images compared to other handcrafted-based models and deep learning-based models.
Although MINT-IQA is designed for assessing the quality of AIGIs, it can also be applied to evaluate the quality of real-captured images.
%other types of images such as Natural Scene Images (NSIs), screen content images, graphic images, etc. 
It can be observed that our proposed method significantly outperforms other models on the AVA \cite{ava} dataset.
 Note that the number of images in the AVA dataset is larger than the other seven datasets, so it may be challenging for other models to learn a broadly representative generic feature in such a large-scale dataset.
Fortunately, the proposed model can learn the feature representation from a more diverse range of image content in large-scale datasets, which leads to a significant improvement.

\vspace{-2mm}
\subsection{Visual Question Answering Acccuracy}

In order to quantitatively measure the capability of our model on explaining human preference for AIGIs. We compare the answers generated by our instruction-tuned model with the other two language models.
% However, the grammar structure of the answer generated by MINT-IQA is quite different from the ground-truth human annotations made in the subjective experiment. The answer generated by MINT-IQA is generally a complete sentence, while the ground-truth human annotations might simply be several choices displayed in the checkboxes or some broken phases instead. 
% Prior metrics used for evaluating the performance of language models generally treat the generated answer and the ground-truth as complete sentences, which do not always hold and bring noise into the evaluation.
However, it is hard to directly evaluate the VQA capability based on sentences. To address this issue, we propose to use visual question answering accuracy ($\text{VQA}_\text{acc}$) for the evaluation of the human preference explanation task, which is computed by directly querying different language models with the same preference-related questions and calculating the accuracy between the model answers and annotations.
 We evaluate the $\text{VQA}_\text{acc}$ mainly based on the five single-choice questions, three of them from the quality perspective, one from the authenticity perspective, and one from the correspondence perspective. 
 % The $\text{VQA}_\text{acc}$ is measured from the three different perspectives separately. 
% The instruction-tuned model is directly instructed with the same setup questions as in the subjective experiment to generate responses, which are subsequently compared to the ground-truth to calculate VQA accuracy.
% Specifically, we make use of Vicuna as the LLM. 
% Given an instruction $I$ with an AI-generated image and its corresponding prompt, the instruction tuned model predicts the human preference answer $\hat{A}$. And 
  % \input{tables/kon}
\begin{table}[!tp]
% \vspace{-11mm}
\renewcommand\arraystretch{1.1}
\caption{VQA accuracy comparison results. The random accuracy is calculated based on the number of options for the single-choice questions. The best results are highlighted in  {\textcolor{red}{RED}}, and the second-best results are highlighted in  {\textcolor{mblue}{blue}}}\label{VQAr}

% \begin{center}
\footnotesize
\centering

\vspace{-2mm}
\begin{tabular}{lccc}\toprule
\textbf{$\text{VQA}_\text{acc}$ } &\textbf{Quality} &\textbf{Authenticity} &\textbf{Correspondence}\\\midrule
Random &25.0 &{\textcolor{mblue}{\textbf{33.3}}} & 33.3 \\
InstructBLIP \cite{instructblip} &8.35 &9.08 & 37.4\\
Llava \cite{liu2023llava} & {\textcolor{mblue}{\textbf{28.3}}} &21.7 &{\textcolor{mblue}{\textbf{45.5}}}\\
% GPT4-V  &49.3 &54.2 &60.5\\
 \hdashline
\rowcolor{gray!20}\textbf{MINT-IQA (Ours)} &\textcolor{red}{\textbf{48.5}} &\textcolor{red}{\textbf{57.3}} & \textcolor{red}{\textbf{59.4}} \\
\rowcolor{gray!20}\textit{Improvement} 
& + 20.2\%
& + 24.0\% 
& + 13.9\%\\
\hline
\end{tabular}
\vspace{-3mm}
% \subtable[Ablation Study on Prompt Segmentation]{
% \begin{tabular}{lrrr}\toprule
% \textbf{Ablation} &\textbf{SRCC} &\textbf{PLCC} &\textbf{KRCC}\\\midrule
% none  &0.8392 &0.8347 & 0.6601 \\
% LLMseg &0.8404 &0.8362 & 0.6619 \\
% \bottomrule
% \end{tabular}}

% \end{center}

\end{table}

Given an AI-generated image and its corresponding prompt, there are $N_q$ quality-relevant questions $Q_\text{q}$ and answers $A_\text{q}$, $N_a$ authenticity-relevant questions $Q_\text{a}$ and their corresponding answers $A_\text{a}$, as well as $N_c$ correspondence-relevant questions $Q_\text{c}$ and their corresponding answers $A_\text{c}$.
% In the following, we do not distinguish between factors from different perspectives including: quality, authenticity and correspondence. However in implementation, we treat them separately to calculate $\text{QAS}_\text{s}$ and $\text{QAS}_\text{c}$.
The $Q_\text{q}$, $Q_\text{a}$ and $Q_\text{c}$ are constructed in the following format:
\vspace{-2mm}
\begin{tcolorbox}[colback=white,colframe=black]
\textit{$Q_\text{q}$ : How is the quality of the image?\\
$Q_\text{a}$ : How is the authenticity of the image?\\
$Q_\text{c}$ : How is the correspondence between the image and its text prompt?}
\end{tcolorbox}
\vspace{-2mm}
By feeding the constructed questions $(Q_\text{q},Q_\text{a},Q_\text{c})$ into a large language model, we can get the human preference answer $\hat{A}$ predicted by the LLM. This process can be expressed as follows:
\begin{equation}
    \hat{A} = \text{LLM}\left(Q_\text{q},Q_\text{a},Q_\text{c}\right).
\end{equation}
Since the answers generated by the model are long sentences, we distill the keywords in the generated answers and match them to the corresponding single-choice questions as follows:
\begin{equation}
    (\hat{A_\text{q}}, \hat{A_\text{a}}, \hat{A_\text{c}}) =\text{Match} ( \text{Distill} (\hat{A}), (Q_\text{q}, Q_\text{a}, Q_\text{c}) ),
\end{equation}
to obtain explicit answers $\hat{A_\text{q}}, \hat{A_\text{a}}, \hat{A_\text{c}}$ for the questions. 
Then, we separately calculate the average accuracy separately among the $N_q$ quality-relevant questions, $N_a$ authenticity-relevant questions and $N_c$ corresponding-relevant questions for an AI-generated image, and obtain the final $\text{VQA}_\text{acc}$ by computing the mean score for all the $K$ images:
\vspace{-2mm}
\begin{equation}
    \text{VQA}_\text{acc\_Quality} = \frac{1}{K}\sum_K\left(\frac{1}{N_q}\sum_{N_q}{\mathbb{I}(\hat{A_\text{q}} = A_\text{q})}\right),
\end{equation}
\begin{equation}
    \text{VQA}_\text{acc\_Authenticity} = \frac{1}{K}\sum_K\left(\frac{1}{N_a}\sum_{N_a}{\mathbb{I}(\hat{A_\text{a}} = A_\text{a})}\right),
\end{equation}
and
\begin{equation}
  \text{VQA}_\text{acc\_Correspondence} = \frac{1}{K}\sum_K\left(\frac{1}{N_c}\sum_{N_c}{\mathbb{I}(\hat{A_\text{c}} = A_\text{c})}\right),
\end{equation}
where $\mathbb{I}(a=b) = 1$ if $a$ equals $b$, and $\mathbb{I}(a=b) = 0$ if $a$ is different from $b$.
In this way,  we can measure the performance of the answers generated by language models and evaluate the preference-related understanding capability of these models on AIGIs. As shown in Table \ref{VQAr}, compared with the random selection accuracy, InstructBLIP \cite{instructblip}, Llava \cite{liu2023llava}, our MINT-IQA model achieves better explanation capability for preference-related questions on AIGIs.  

\vspace{-2mm}

% \subsection{Loss Function}

\begin{table*}[th]
\begin{center}
% \vspace{-3mm}
\small

% \tablestyle{2pt}{1}
\renewcommand*{\arraystretch}{0.9}
% \vspace{-5mm}
\caption{SRCC results of cross-dataset verification from the \textbf{\underline{quality}} perspective. \textbf{Bold} entries
indicate the best performers.}
\vspace{-3mm}
\scalebox{1.1}{\begin{tabular}{c|cc|cc|cc|cc}
\hline
{\fontsize{7.6}{8}\selectfont Train on} & \multicolumn{2}{c|}{{\fontsize{7.6}{8}\selectfont KonIQ \cite{hosu2020koniq}}} & \multicolumn{2}{c|}{{\fontsize{7.6}{8}\selectfont SPAQ \cite{fang2020perceptual}}} & \multicolumn{2}{c|}{{\fontsize{7.6}{8}\selectfont AGIQA-3K \cite{database/agiqa} (quality)}} &\multicolumn{2}{c}{{\fontsize{7.6}{8}\selectfont AIGCIQA2023+ (quality)}}\tabularnewline
\hline
{\fontsize{7.6}{8}\selectfont Test on} & {\fontsize{7.6}{8}\selectfont SPAQ } & {\fontsize{7.6}{8}\selectfont KonIQ} & {\fontsize{7.6}{8}\selectfont KonIQ }& {\fontsize{7.6}{8}\selectfont SPAQ }& {\fontsize{7.6}{8}\selectfont AIGCIQA2023+}&{\fontsize{7.6}{8}\selectfont  AGIQA-3K}&{\fontsize{7.6}{8}\selectfont AGIQA-3K}&{\fontsize{7.6}{8}\selectfont AIGCIQA2023+}\tabularnewline
\hline
{\fontsize{7.6}{8}\selectfont TReS \cite{golestaneh2022no}} & {\fontsize{7.6}{8}\selectfont0.765} & {\fontsize{7.6}{8}\selectfont 0.907} & {\fontsize{7.6}{8}\selectfont0.828} & {\fontsize{7.6}{8}\selectfont0.915} & {\fontsize{7.6}{8}\selectfont0.747} & {\fontsize{7.6}{8}\selectfont0.646} & {\fontsize{7.6}{8}\selectfont0.634} &{\fontsize{7.6}{8}\selectfont 0.741}\tabularnewline
{\fontsize{7.6}{8}\selectfont  \textbf{MINT-IQA (Ours)}} & {\fontsize{7.6}{8}\selectfont \textbf{0.772}} &{\fontsize{7.6}{8}\selectfont\textbf{0.927}} & {\fontsize{7.6}{8}\selectfont\textbf{0.892}} & {\fontsize{7.6}{8}\selectfont\textbf{0.927}} & {\fontsize{7.6}{8}\selectfont\textbf{0.817}}& {\fontsize{7.6}{8}\selectfont \textbf{0.892}}& {\fontsize{7.6}{8}\selectfont \textbf{0.770}}& {\fontsize{7.6}{8}\selectfont \textbf{0.880}}\tabularnewline
\hline
\end{tabular}\label{crossdata}}
% \vspace{-0.8em}
\end{center}
\vspace{-6mm}
\label{cross}
\end{table*}

\subsection{Cross-Dataset Evaluation}

We further conduct an experiment to test the generalization ability of MINT-IQA in a more challenging cross-dataset setting. Specifically, we train the models on the training set of one database and directly test the generalization ability on the test set of another database.
The cross-dataset evaluation experiment is conducted on both the AIGC IQA databases and traditional IQA databases. As shown in Table \ref{crossdata}, our proposed model MINT-IQA achieves better results compared to latest models TReS \cite{golestaneh2022no} for all tasks.
\vspace{-2mm}
\subsection{Ablation Study}

To validate the contributions of the different modules in MINT-IQA, we further conduct ablation studies. The results are demonstrated in Tables \ref{tab:9a}-\ref{tab:9e}.  We report the averaged SRCC, PLCC, KRCC scores from three perspectives on our AIGCIQA2023+ database for each ablation experiment.
\subsubsection{Comparison between Different Model Backbone}
 To verify the rationality and effectiveness of the Q-Former in the proposed MINT-IQA model, we change the model backbone to CLIP \cite{radford2021learning} and BLIP \cite{li2022blip} for comparison. We directly add our multi-head quality regression to CLIP \cite{radford2021learning} and BLIP \cite{li2022blip}, and train them in a similar way. Table \ref{tab:9a} manifests that our Q-Former based method outperforms conventional CLIP \cite{radford2021learning} and BLIP \cite{li2022blip} as the model backbone.
\subsubsection{Comparison between Different Loss Functions}
To further validate the effectiveness of the adopted loss function, we compare it with the cross-entropy loss. 
We can observe from Table \ref{tab:9b} that the L1 loss is more effective in dealing with our regression-based IQA tasks.
\subsubsection{Comparison between Different Fix Rates}
The training efficiency of a model is crucial in practical applications. Generally, models with fewer active parameters may have better training efficiency but worse performance results. Since the image encoder accounts for the largest proportion of parameters in the model, we conduct ablation experiments to fix some layers of the image encoder.
From Table \ref{tab:9c} we can observe that freezing the whole model leads to the worst performance, while the fix rate of 0.7 leads to the best performance. It is reasonable since the feature extraction model needs to be trained for this specific task, but due to the dataset size limitation, training the whole model may lead to the overfitting problem.
It is also beneficial to consider the balance between the performance and computational efficiency.
% The model with a low fix rate has too many parameters leading to rapid convergence and consequent overfitting, which harms model performance.

\subsubsection{Contribution of LLM Segmentation and the Feedback Reward Part}
To verify the effectiveness of the design of the two improvement parts for the performance, we launch further ablation experiments of adding the LLM segmentation module and feedback reward module, and show the results in Table \ref{tab:9d}. Using the prompt after LLM Segmentation as the input leads to better performance than raw prompt. The reward feedback part can also further improve the performance of our model.
Moreover, the combination of two further improved modules achieves better results compared to only using one part. This could suggest that more granular input processing and dynamic adjustment of evaluation metrics based on explanatory feedback can lead to a more accurate alignment with human preference judgement.\begin{table}[!tp]
% \vspace{-8mm}
% \caption{Ablation study results on AIGCIQA2023 dataset. We report the average SRCC, PLCC, KRCC scores from three perspectives for each ablation experiment. The VQA accuracy is also reported. We \textbf{bold} the best results in each table.}\label{tab:ablation}
% \vspace{-2mm}
% \begin{center}
\footnotesize
\centering
\renewcommand\arraystretch{1.1}
\caption{Ablation study on model backbone.}\label{tab:9a}
\vspace{-2mm}

\scalebox{1}{
\begin{tabular}{lrrr}
\toprule
\textbf{Backbone} &\textbf{SRCC} &\textbf{PLCC} &\textbf{KRCC}\\\midrule
CLIP \cite{radford2021learning} &0.7086 &0.7003 & 0.5330 \\
% CLIP &0.8392 &0.8347 & 0.6601 \\
BLIP \cite{li2022blip}&0.7937 &0.7860 & 0.6038 \\
Q-Former &\textbf{0.8419} &\textbf{0.8351} & \textbf{0.6432}\\
\bottomrule
\end{tabular}}
\vspace{-2mm}
\end{table}

\begin{table}[!tp]
\footnotesize
\centering
\caption{Ablation study on loss function.}\label{tab:9b}
\vspace{-2mm}
\scalebox{0.9}{
\begin{tabular}{lrrr}\toprule
\textbf{Loss Function} &\textbf{SRCC} &\textbf{PLCC} &\textbf{KRCC}\\\midrule
Cross-entropy &0.7312 &0.7199 & 0.5244\\
L1 loss &\textbf{0.8419}&\textbf{0.8351} & \textbf{0.6432} \\
\bottomrule
\end{tabular}}
\vspace{-2mm}
\end{table}

\begin{table}[!tp]
\footnotesize
\centering
\caption{Ablation study on the fix rate of image encoder.}\label{tab:9c}
\vspace{-2mm}
\scalebox{1}{
\begin{tabular}{lcccr}\toprule
\textbf{Fix Rate} &\textbf{SRCC} &\textbf{PLCC} &\textbf{KRCC} &\textbf{GPU Memory}\\\midrule
1.0 &0.8235 &0.8190 & 0.6229 & \textbf{9450M}\\
0.8 &0.8374 &0.8325 & 0.6372 & 14846M \\
0.7 &\textbf{0.8419}&\textbf{0.8351} &\textbf{0.6432}& 17586M \\
0.6 &0.8411 &0.8346 & 0.6417 & 20336M\\
0.5 &0.8403 &0.8338 & 0.6398 & 22330M \\
% 0.4 &0.8381 &0.8315 & 0.6609 & \textbf{41604M}\\
\bottomrule
\end{tabular}}
\vspace{-2mm}
\end{table}

% \begin{table}[!tp]
% \footnotesize
% \centering
% \caption{Ablation study on LLM segmentation}\label{tab:9d}
% \vspace{-2mm}
% \begin{tabular}{lrrr}\toprule
% \textbf{Segmentation} &\textbf{SRCC} &\textbf{PLCC} &\textbf{KRCC}\\\midrule
% without &0.8392 &0.8347 & 0.6601 \\
% with &\textbf{0.8404} &\textbf{0.8362} & \textbf{0.6619} \\
% \bottomrule 
% \end{tabular}
% \vspace{-2mm}
% \end{table}

% \begin{table}[!tp]
% \footnotesize
% \centering
% \caption{Ablation study on the reward part of the model.}\label{tab:9f}
% \vspace{-2mm}
% \begin{tabular}{lrrr}\toprule
% \textbf{Reward} &\textbf{SRCC} &\textbf{PLCC} &\textbf{KRCC}\\\midrule
% without &0.8392 &0.8347 & 0.6601 \\
% with &\textbf{0.8508} &\textbf{0.8456} & \textbf{0.6728} \\
% \bottomrule 
% \end{tabular}
% \vspace{-2mm}
% \end{table}

\begin{table}[!tp]
\footnotesize
\centering
\caption{Improvements of LLM segmentation and feedback reward.}\label{tab:9d}
\vspace{-2mm}\scalebox{1}{
\begin{tabular}{lrrr}\toprule
\textbf{Segmentation} &\textbf{SRCC} &\textbf{PLCC} &\textbf{KRCC}\\\midrule
MINT-IQA (Ours) &0.8419 &0.8351 & 0.6432 \\
 % + LoRA & 0.8427 & 0.8357 & 0.6328 \\
 + LLM Seg. &0.8456 &0.8400& 0.6482 \\
 + Reward &0.8499 &0.8447 &0.6539\\
 + LLM Seg. + Reward &\textbf{0.8510} &\textbf{0.8453} &\textbf{0.6546}\\
\bottomrule 
\end{tabular}}
\vspace{-2mm}
\end{table}

\begin{table}[!tp]
\footnotesize
\centering
\caption{Ablation study on training strategy.}\label{tab:9e}
\vspace{-2mm}
\scalebox{0.9}{
\begin{tabular}{llrrr}\toprule
\textbf{Dimension} &\textbf{Training Strategy} &\textbf{SRCC} &\textbf{PLCC} &\textbf{KRCC}\\\midrule
\multirow{3}{*}{Quality} &MINT-IQA-joint &0.8801 &0.8870 & 0.6841\\
 &MINT-IQA-LoRA &0.8841&0.8866 & 0.6888\\
&MINT-IQA-single &\textbf{0.8871}&\textbf{0.8948} & \textbf{0.6946} \\
\midrule
\multirow{3}{*}{Authenticity} &MINT-IQA-joint &0.8229 &0.8127 & 0.6223\\
 &MINT-IQA-LoRA  &0.8286 &0.8137 & 0.6264\\
&MINT-IQA-single &\textbf{0.8341}&\textbf{0.8260} & \textbf{0.6373} \\
\midrule
\multirow{3}{*}{Correspondence} &MINT-IQA-joint &0.8226 &0.8055 & 0.6231\\
 &MINT-IQA-LoRA &0.8154 &0.8067 & 0.6133\\
&MINT-IQA-single &\textbf{0.8253}&\textbf{0.8117} & \textbf{0.6268} \\
\bottomrule
\end{tabular}}
\vspace{-2mm}
\end{table}

\begin{table}[!tp]
\footnotesize
\centering
\caption{Ablation study on zero convolution connection. VQA accuracy is from the test set of 480 images and 2400 detailed explanations.}\label{tab:9f}
\vspace{-2mm}\scalebox{1}{
\begin{tabular}{lccc}\toprule
\textbf{Zero Conv} &\textbf{Quality} &\textbf{Authenticity} &\textbf{Correspondence}\\\midrule
without &47.2 &54.4 & 57.1 \\
with &\textbf{48.5} &\textbf{57.3} & \textbf{59.4}\\
\bottomrule 
\end{tabular}}
\vspace{-2mm}
\end{table}

\subsubsection{Comparison of Different Training Strategies}
\begin{figure}[t]
% \vspace{-5mm}
	\centering
	\includegraphics[width=\linewidth]{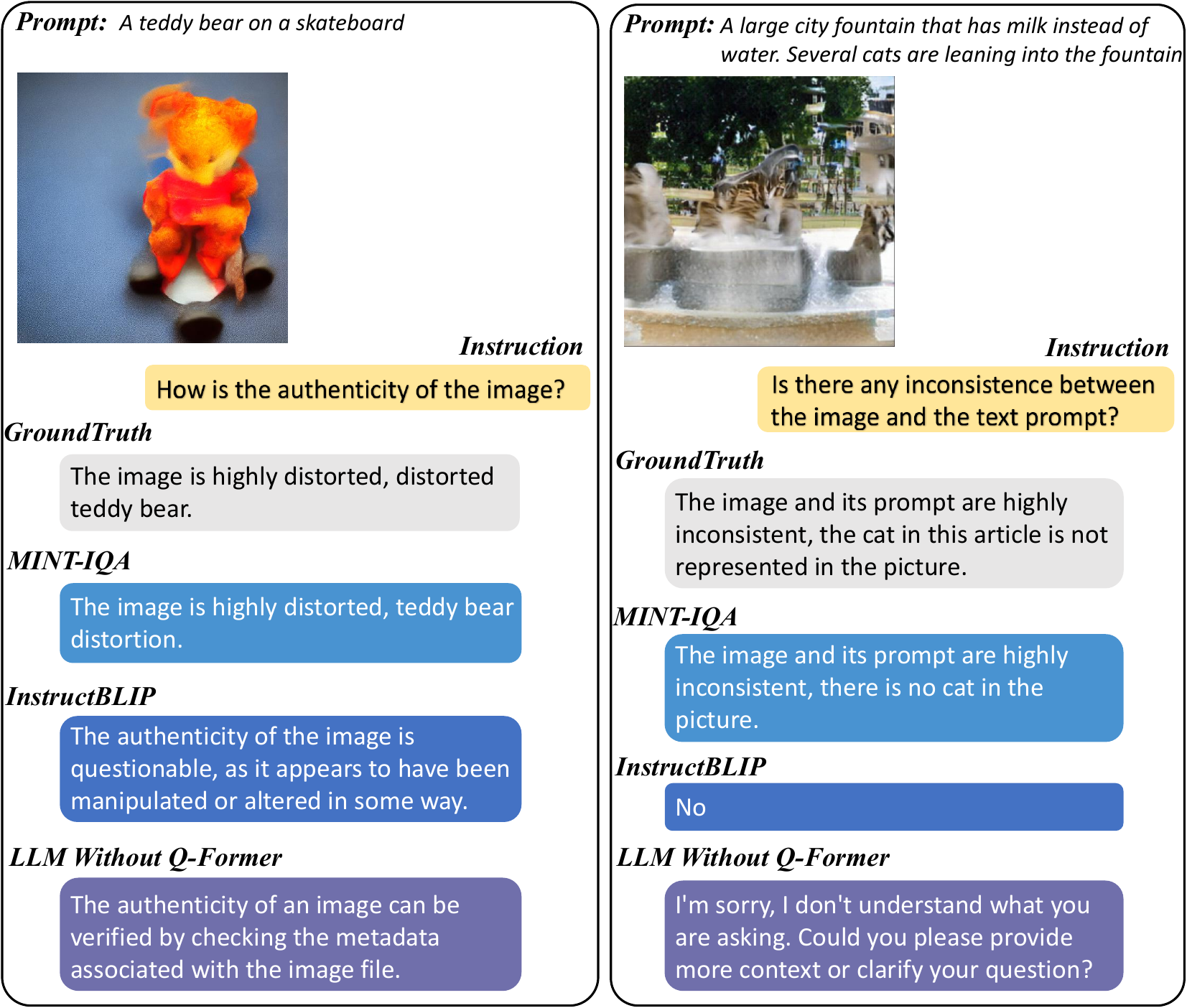}
 \vspace{-5mm}
	\caption{Ablation study on the vision-language instruction tuning strategy. With fine-tuning on AIGCIQA2023+ database, our proposed model achieves better performance on visual question answering.}
	\label{ablation}
 \vspace{-4mm}
\end{figure}
We explore the impact of the fine-tuning strategy on the training of the MINT-IQA model, as shown in Table \ref{tab:9e}. Specifically, we utilize Low-Rank Adaptation (LoRA) setting the rank parameter to $r=16$. LoRA enables efficient model adaptation by adding a limited number of trainable parameters, which reduces computational costs while preserving performance. Our experimental results demonstrate that incorporating LoRA fine-tuning leads to a modest yet statistically significant improvement in the model’s SRCC. 
We further explore two distinct training methodologies: MINT-IQA-joint and MINT-IQA-single. The joint training approach simultaneously optimizes three regression losses within a single backward pass, resulting in a unified set of weights that allow for the concurrent inference of three quality scores. This method offers computational efficiency by maintaining only one set of weights. In contrast, the sequential training approach, referred to as MINT-IQA-single, independently optimizes each regression loss through separate backward passes, thereby generating three distinct sets of weights tailored to each specific quality score. Although MINT-IQA-single requires three separate training iterations, it achieves higher SRCC values compared to MINT-IQA-joint, indicating superior alignment with human perceptual judgments. This highlights a trade-off between computational efficiency and assessment accuracy, suggesting that while joint training is more resource-efficient, sequential training provides enhanced performance in quality assessment.
% We can observe that the model performance gained by the single-training strategy is better than the joint-training strategy, which indicates that the three evaluation perspectives have a certain correlation and may have mutual influence during the joint-training process.
\subsubsection{Contribution of zero convolution}
Table \ref{tab:9f} validates the significance of initializing the connection part between the two Q-Formers with zero convolution.  
The removal of the connection between two Q-Formers through zero convolution leads to poorer performance. This highlights the effectiveness of inserting the scoring features into the explanation module for better VQA performance.
% The removal of the connection between the two modules leads to worse performance on VQA accuracy.

\subsubsection{Contribution of the vision-language instruction tuning strategy}
Finally, we investigate the impact of the vision-language instruction tuning strategy, and show the results in Fig. \ref{ablation}. The answers generated without fine-tuning on our AIGCIQA2023+ database show poor consistency with human preference. If we further remove the connection between the Q-Former and the LLM, the answer performance drop is much more severe. This further highlights the value of vision-language instruction tuning strategy in aligning AI-generated responses with human preferences. 

% Additionally, the fix rate of the image encoder (Table \ref{tab:9c}) and the removal of LLM segmentation (Table \ref{tab:9d}) also influence the model performance. Furthermore, we prove the great significance of the connection part between the two Q-Formers through ablation study on the zero convolution (Table \ref{tab:9e}) and reward part (Table \ref{tab:9d}). 

%with no instruction-aware representations, the answer performance drop is much more severe.
% Additionally, if we further remove the Q-Former and merely give instructions to the LLM with no instruction-aware representations, the answer performance drop is much more severe.
% The factors in prompt segmentation, (loss function ): pair loss. The result show that removing any single factor leads to performance degradation, which confirms that they all contribute to the performance results in Tab.\ref{tab:AIGCIQA2023-results}.
% \vspace{-10mm}

\section{Conclusion }
% \vspace{-1mm}
This work aims to better understand and evaluate human visual preferences for AIGIs. 
We extend the AIGCIQA2023 database to AIGCIQA2023+, which includes more fine-grained preference labels from the perspectives of quality, authenticity, and text-image correspondence. The experimental analysis demonstrates that these three dimensions can reflect different aspects of human visual preferences on AIGIs, which further manifests the evaluation of Quality of Experience (QoE) for AIGIs should be considered from multiple dimensions. Based on the constructed database, we propose a novel method termed MINT-IQA, which allows for a more comprehensive evaluation of the human preferences for AIGIs from multiple perspectives, and gives fine-grained explanations for the preference-related questions.
Extensive experimental results demonstrate the effectiveness of MINT-IQA on both preference evaluation and explanation for AIGIs.

\section{Limitation and Future Works}
The current MINT-IQA approach primarily depends the training on subjective human ratings to achieve good consistency with human ratings. While this method provides valuable insights into perceived alignment quality, it is time-consuming and may lack scalability. It is valuable to explore the integration of automated detection mechanisms and high-level semantic analysis to assess image-text alignment in future works. By leveraging advanced techniques such as object detection, scene understanding, and semantic consistency checks, the model can achieve more objective and efficient evaluations, reducing dependence on manual annotations. As multi-modal perception becomes increasingly prevalent, many multi-modal generation works have also been proposed. It is deserved to develop novel fusion strategies that can effectively combine visual, auditory, video, mesh, and potentially other sensory data to perform quality assessment for multi-modal generations, enabling the assessment of diverse media types within a unified framework.

\section{Acknowledgement}
This work was supported in part by the National Key R\&D Program of China under Grant 2021YFE0206700, in part by the National Natural Science Foundation of China under Grants 62401365, 62271312, 62225112, 62132006, and in part by the Shanghai Pujiang Program under Grant 22PJ1407400.

% {
\bibliographystyle{IEEEtran}

\bibliography{main.bib}
% }
\newpage
\section{Biography Section}
\begin{IEEEbiography}[{\includegraphics[width=1in,height=1.25in,clip,keepaspectratio]{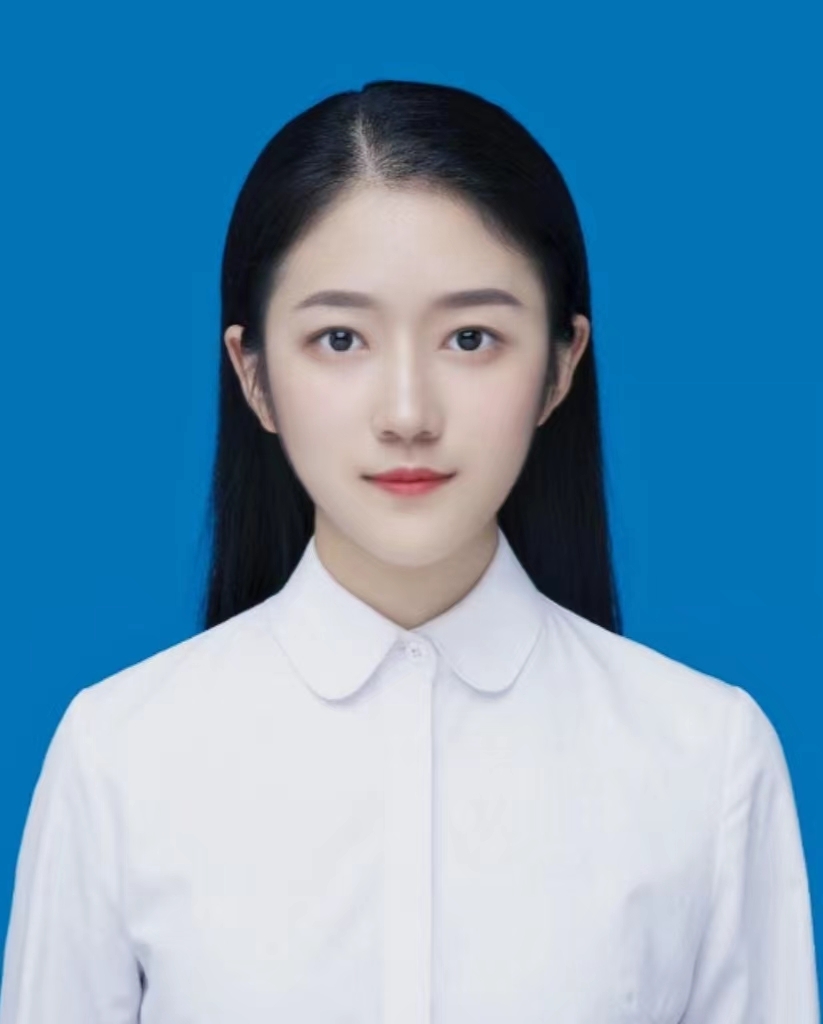}}]{Jiarui Wang}
received the B.E. degree from Shanghai Jiao Tong University, Shanghai, China, in 2024. She is currently working toward the Ph.D. degree with the Department of Electronic Engineering, Shanghai Jiao Tong University. Her research interests include perceptual quality assessment, quality of experience and multimodal signal processing.
\end{IEEEbiography}
\begin{IEEEbiography}[{\includegraphics[width=1in,height=1.25in,clip,keepaspectratio]{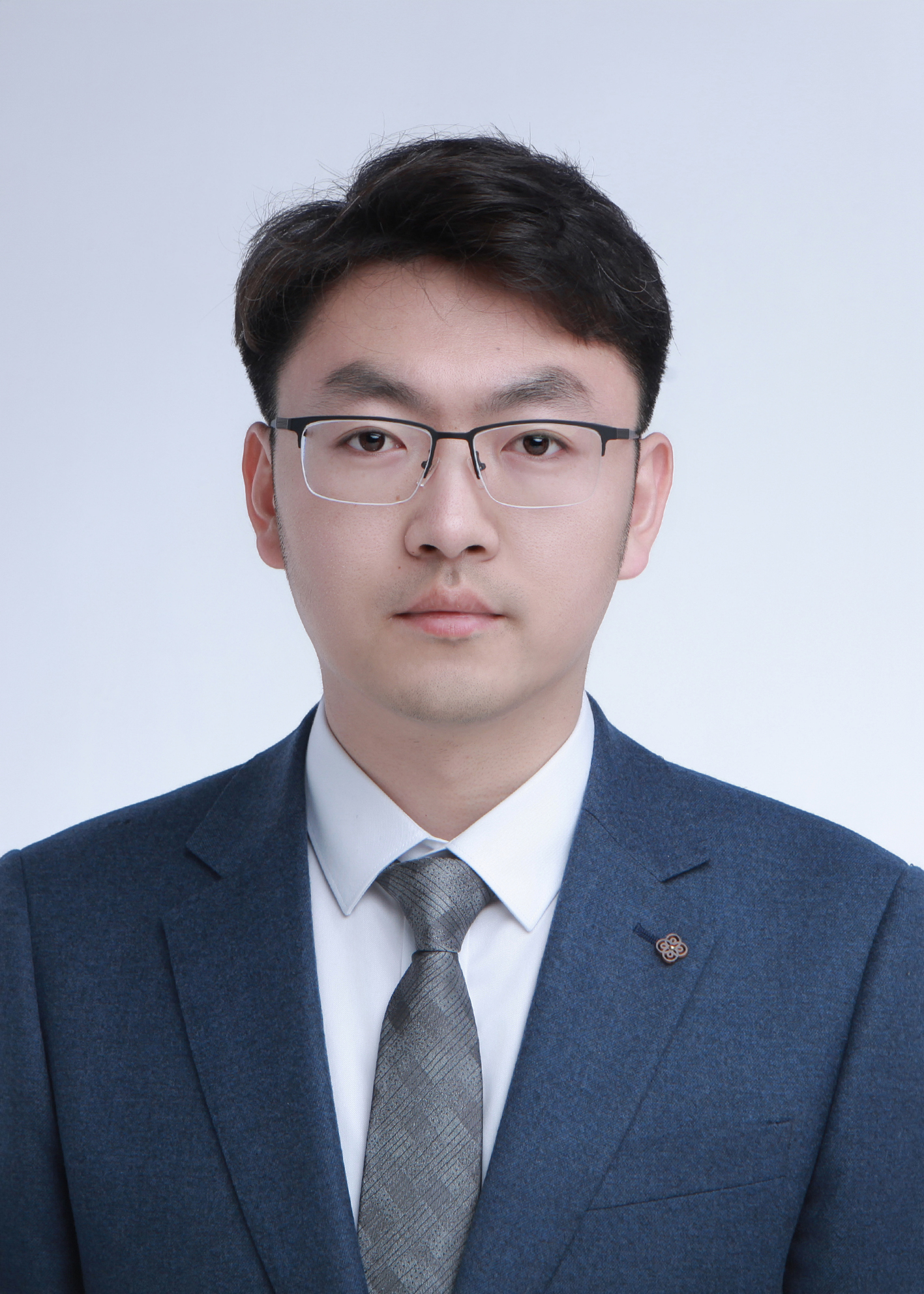}}]{Huiyu Duan}
received the B.E. degree from the University of Electronic Science and Technology of China, Chengdu, China, in 2017, and the Ph.D. degree from Shanghai Jiao Tong University, Shanghai, China, in 2024, where he is currently a Postdoc with the Institute of Image Communication and Network Engineering. From Sept. 2019 to Sept. 2020, he was a visiting Ph.D. student at the Schepens Eye Research Institute, Harvard Medical School, Boston, USA. He received the Best Paper Award of IEEE International Symposium on Broadband Multimedia Systems and Broadcasting (BMSB) in 2022. His research interests include perceptual quality assessment, quality of experience, visual attention modeling, extended reality (XR), and multimedia signal processing.
\end{IEEEbiography}
\begin{IEEEbiography}[{\includegraphics[width=1in,height=1.25in,clip,keepaspectratio]{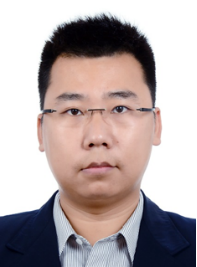}}]{Guangtao Zhai}
 (SM’19) received the B.E. and M.E. degrees from Shandong University, Shandong, China, in 2001 and 2004, respectively, and the Ph.D. degree from Shanghai Jiao Tong University, Shanghai, China, in 2009, where he is currently a Research Professor with the Institute of Image Communication and Information Processing. From 2008 to 2009, he was a Visiting Student with the Department of Electrical and Computer Engineering, McMaster University, Hamilton, ON, Canada, where he was a Post-Doctoral Fellow from 2010 to 2012. From 2012 to 2013, he was a Humboldt Research Fellow with the Institute of Multimedia Communication and Signal Processing, Friedrich Alexander University of Erlangen-Nuremberg, Germany. He received the Award of National Excellent Ph.D. Thesis from the Ministry of Education of China in 2012. His research interests include multimedia signal processing and perceptual signal processing.
\end{IEEEbiography}
\begin{IEEEbiography}[{\includegraphics[width=1in,height=1.25in,clip,keepaspectratio]{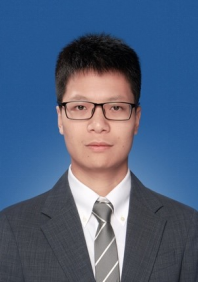}}]{Xiongkuo Min}
received the B.E. degree from Wuhan University, Wuhan, China, in 2013, and the Ph.D. degree from Shanghai Jiao Tong University, Shanghai, China, in 2018, where he is currently a tenure-track Associate Professor with the Institute of Image Communication and Network Engineering. From Jan. 2016 to Jan. 2017, he was a visiting student at University of Waterloo. From Jun. 2018 to Sept. 2021, he was a Postdoc at Shanghai Jiao Tong University. From Jan. 2019 to Jan. 2021, he was a visiting Postdoc at The University of Texas at Austin. He received the Best Paper Runner-up Award of IEEE Transactions on Multimedia in 2021, the Best Student Paper Award of IEEE International Conference on Multimedia and Expo (ICME) in 2016, and the excellent Ph.D. thesis award from the Chinese Institute of Electronics (CIE) in 2020. His research interests include image/video/audio quality assessment, quality of experience, visual attention modeling, extended reality, and multimodal signal processing.
\end{IEEEbiography}
\vfill
\end{document}